\documentclass[11pt,twoside,openright]{report}

\usepackage[a4paper,top=2.6cm,bottom=2.7cm,margin=2.6cm]{geometry}

\usepackage{ifxetex}
\ifxetex
  \usepackage{mathspec}
  \usepackage{fontspec}
  \defaultfontfeatures{Scale=MatchLowercase}

(Digits,Greek,Latin)[Numbers=OldStyle]{Linux Libertine}
\else
  \usepackage{libertine}
  \usepackage{inconsolata}
\fi

\makeatletter
\def\mdseries@tt{m}
\makeatother

\usepackage{microtype}

\usepackage{xcolor}
\definecolor{navy}{HTML}{102573}

\PassOptionsToPackage{
  colorlinks=true,
  linkcolor=navy,
  urlcolor=navy,
  citecolor=navy
}{hyperref}

\usepackage{fancyref}
\usepackage{url}

\usepackage{apacite}

\usepackage{enumitem}
\setlist{
  listparindent=\parindent,
  parsep=0pt,
}

\usepackage{titling}
\usepackage[onehalfspacing]{setspace}

\usepackage{emptypage}

\usepackage{booktabs}
\usepackage{multirow}

\usepackage{graphicx}
\usepackage{float}
\usepackage{subcaption}
\usepackage[labelfont=bf, 
            textfont=it]{caption}

\usepackage{todonotes}

\definecolor{bg}{rgb}{0.95,0.95,0.95}
\usepackage{listing}
\usepackage[frozencache]{minted}
\setminted{bgcolor=bg,fontsize=\small}

\usepackage{titlesec}
\titleformat{\chapter}[display]
{\normalfont\huge\bfseries}{\chaptertitlename\ \thechapter}{20pt}{\Huge}
\titlespacing*{\chapter}{0pt}{-30pt}{30pt}

\newenvironment{longlisting}{\captionsetup{type=listing}}{}

\newcommand*{\fancyreflstlabelprefix}{lst}

\fancyrefaddcaptions{english}{%
  \providecommand*{\freflstname}{listing}%
  \providecommand*{\Freflstname}{Listing}%
}

\frefformat{plain}{\fancyreflstlabelprefix}{\freflstname\fancyrefdefaultspacing#1}
\Frefformat{plain}{\fancyreflstlabelprefix}{\Freflstname\fancyrefdefaultspacing#1}

\frefformat{vario}{\fancyreflstlabelprefix}{%
  \freflstname\fancyrefdefaultspacing#1#3%
}
\Frefformat{vario}{\fancyreflstlabelprefix}{%
  \Freflstname\fancyrefdefaultspacing#1#3%
}

\newcommand*{\fancyrefapplabelprefix}{app}

\fancyrefaddcaptions{english}{%
  \providecommand*{\frefappname}{appendix}%
  \providecommand*{\Frefappname}{Appendix}%
}

\frefformat{plain}{\fancyrefapplabelprefix}{\frefappname\fancyrefdefaultspacing#1}
\Frefformat{plain}{\fancyrefapplabelprefix}{\Frefappname\fancyrefdefaultspacing#1}

\frefformat{vario}{\fancyrefapplabelprefix}{%
  \frefappname\fancyrefdefaultspacing#1#3%
}
\Frefformat{vario}{\fancyrefapplabelprefix}{%
  \Frefappname\fancyrefdefaultspacing#1#3%
}

\title{Reducing Noise from Competing Neighbours:
Word~Retrieval with~Lateral Inhibition in~Multilink}
\author{Aaron van Geffen $\langle3058026\rangle$}
\date{November 6, 2019}

\begin{document}
    \pagenumbering{gobble}
    \pagestyle{empty}

    \begin{titlepage}
	\begin{center}
		\textsc{\LARGE Master's Thesis\\Artificial Intelligence}\\[1.5cm]
		\includegraphics[height=100pt]{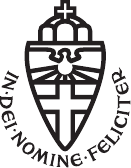}

		\vspace{0.4cm}
		\textsc{\Large Radboud University}\\[1cm]
		\hrule
		\vspace{0.4cm}
		\begin{spacing}{2.0}
		\textbf{\huge \thetitle }
		\end{spacing}
		\vspace{0.1cm}
		\hrule
		\vspace{2cm}
		\begin{minipage}[t]{0.45\textwidth}
			\begin{flushleft} \large
				\textit{Author:}\\
				Aaron van Geffen\\
				3058026
			\end{flushleft}
		\end{minipage}
		\begin{minipage}[t]{0.45\textwidth}
			\begin{flushright} \large
				\textit{First supervisor/assessor:}\\
				prof. dr. Ton Dijkstra\\
				\texttt{t.dijkstra@donders.ru.nl}\\[1.3cm]

				\textit{Second assessor:}\\
				dr. Frank L\'eon\'e \\
				\texttt{f.leone@donders.ru.nl}
			\end{flushright}
		\end{minipage}
		\vfill
		{\large \thedate}
	\end{center}
\end{titlepage}

    \cleardoublepage

    \pagenumbering{roman}
    \pagestyle{plain}

    \pdfbookmark[0]{Abstract}{abstract}
\chapter*{Abstract}

Multilink is a computational model for word retrieval in monolingual and multilingual individuals
under different task circumstances \cite{dijkstra2018multilink}.
In the present study, we added lateral inhibition to Multilink's lexical network.
Parameters were fit on the basis of reaction times from the English, British, and Dutch Lexicon Projects.
We found a maximum correlation of 0.643 (N=1,205) on these data sets as a whole.
Furthermore, the simulations themselves became \emph{faster} as a result of adding lateral inhibition.
We tested the fitted model to stimuli from a neighbourhood study \shortcite{mulder2018}.
Lateral inhibition was found to improve Multilink's correlations for this study,
yielding an overall correlation of 0.67.

Next, we explored the role of lateral inhibition as part of the model's task/decision system
by running simulations on data from two studies concerning interlingual homographs \shortcite{vanlangendonckIP, goertz2018}.
We found that, while lateral inhibition plays a substantial part in the word selection process,
this alone is not enough to result in a correct response selection.
To solve this problem, we added a new task component to Multilink, especially designed
to account for the translation process of interlingual homographs, cognates, and language-specific control words.
The subsequent simulation results showed patterns remarkably similar to those in the \citeauthor{goertz2018} study.
The isomorphicity of the simulated data to the empirical data was further attested by an
overall correlation of 0.538 (N=254) between reaction times and simulated model cycle times,
as well as a condition pattern correlation of 0.853 (N=8).

We conclude that Multilink yields an excellent fit to empirical data, particularly when a task-specific setting
of the inhibition parameters is allowed.


    \clearpage
\pdfbookmark[0]{Acknowledgements}{acknowledgements}
\chapter*{Acknowledgements}

During my work on this thesis, I have been fortunate to be supported and inspired by many people.
Before we dive into the theoretical matter, I would like to take the opportunity to express my gratitude
to them.

First and foremost, I would like to thank Ton Dijkstra, who inspired and supervised this thesis.
Over the course of my internship and the subsequent writing of this thesis,
I have come to admire his working knowledge of the field.
Moreover, I am very appreciative of his mentorship, both on academia and life in general.
I~could not have wished for a more enthusiastic and committed supervisor.

Frank L\'eon\'e, for acting as a second assessor to the process.
Notably, I am grateful for him virtually attending my thesis presentation using Skype
when physical presence turned out not to be possible.

Randi Goertz, for in-depth discussions about the model around the project's inception.
These conversations have certainly helped shape my internship and the resulting thesis.

Koji Miwa, for inviting me to present some preliminary results to this thesis
in his lab at Nagoya University, Japan.
I particularly like this aspect of academia, sharing and indeed fostering of knowledge,
and hope to have sparked some interest into (computational) modelling in the attendees.

James McQueen, whose course on \emph{Word Recognition} introduced me to modelling work
in the field of computational psycholinguistics. His work on the Shortlist model in particular
has been very formative to the work on lexical competition presented in this thesis.

Makiko Sadakata, for introducing me to the renewed AI master programme in Nijmegen
while I had been focusing on doing a master's in Japan.
I am very happy and grateful I enrolled in the programme.

Johanna de Vos,
Arushi Garg,
Austin Howard,
Marc Schoolderman,
Ted Thurlings, and
Willem de Wit
-- thanks for all the spontaneous cups of tea, good conversation, better advice, and great friendship.

Margot Mangnus,
Garima Kar,
Laura Toron, and
Janna Schulze
-- thank you for making the DCC internship room a livelier place.

Haruna Chinzei, thank you for coming into my life during the earlier stages of this thesis.
While I am leaving my thesis work behind, I am glad to continue to have you in my life.

Finally, I would like to thank my parents, Joop and Loes van Geffen, for their love and support.


    \clearpage
    \pdfbookmark[0]{\contentsname}{toc}
    \tableofcontents


    \cleardoublepage

    \setcounter{page}{1}
    \pagenumbering{arabic}
    \pagestyle{plain}

    \chapter{Introduction}

Words are the building blocks of the sentences we use in our everyday communication.
Hence, they are the units of language that most psycholinguistic research focuses on \cite{harley2014psychology}.
The \emph{monolingual} processes of retrieving words during comprehension and production have been thoroughly investigated during the past few decades,
and are generally well understood.
However, there is no such general consensus regarding word retrieval processes in people who speak more than one language.
The most complicated process involving word retrieval in such bilinguals and multilinguals is probably the \emph{word translation} process,
as it involves comprehension, semantic processing, and production, all nearly at the same time.

Experimental studies have shown that some words are easier to translate than others.
A special class of such words, translation equivalents with considerable overlap in form, are called \emph{cognates}.
This cross-linguistic overlap can concern orthography or phonology, or both.
For example, the word `tunnel' shares both its form and meaning between Dutch and English.
In experimental tasks, participants have been found to process cognates faster
and with fewer errors than in control conditions with matched one-language words \cite{christoffels2007}.
This performance difference is called the \emph{cognate facilitation effect}.

However, some words share the same orthography between languages, but unlike cognates, lack any semantic overlap.
These are called \emph{interlingual homographs}, colloquially known as \emph{false friends}.
For example, the word `room' in Dutch translates to the English word `cream',
while the English word `room' translates to the Dutch word `kamer'.
Such words may be more \emph{difficult} to translate, as the two readings of the item may compete.
The selection of the correct reading of the item thus requires an \emph{inhibition} of the other reading.

In order to to better understand the mechanisms underlying the word translation process,
the scientific theories pertaining to these mechanisms can be implemented in a \emph{computational model}.
This allows us to consistently test our hypotheses by presenting word stimuli to the model,
and comparing its simulation results to what we find in empirical data.
If the simulations yield result patterns comparable to those in the experiments (assessed by model-to-data
comparison), the model's workings may be considered as isomorphic to the human word retrieval process and therefore an adequate
representation of this subdomain of reality.

There is not one clear-cut approach to modelling, however.
Many modern approaches to neural networks define the network \emph{structure} in terms of capacities and links,
but not the \emph{function} of those nodes.
Instead, these functions are \emph{trained} in a process commonly referred to as \emph{machine learning}.
The localist-connectionist method \cite{page2000} approaches the issue differently.
In the first method, weights for connections and meaning for nodes in the network are assigned
through a computationally intensive learning process,
while in the second method these weights and meanings are assigned by the experimenter.
Both methods have their advantages and disadvantages.
However, as we will see in the next chapters, these localist-connectionist models
provide a powerful theoretical account for empirical data.

In this thesis, we will investigate several extensions to the localist-connectionist model \emph{Multilink}
to better account for translation processes.
Let us start by discussing the model as it is presented in \shortciteA{dijkstra2018multilink}.

\section{The Multilink model}

\begin{figure}[!t]
    \centering
    \includegraphics[width=0.8\linewidth]{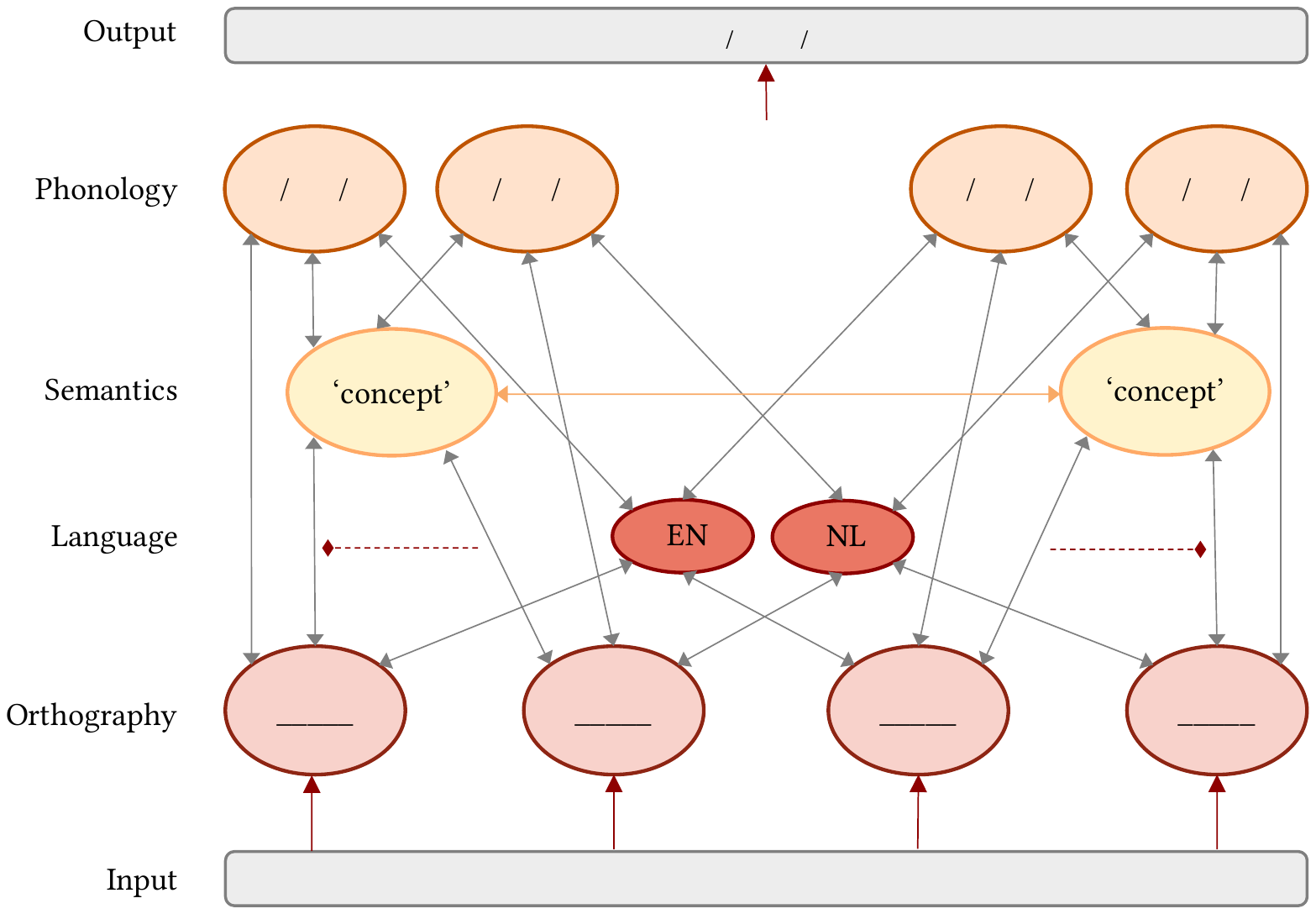}
    \caption{The architecture of Multilink's lexical network, illustrating the different kinds of
        representational nodes and their connections.}
    \label{fig:multilink}
\end{figure}

The Multilink model \cite{dijkstra2018multilink, dijkstra2010towards} is a localist-connectionist model for
monolingual and bilingual word recognition and word translation.
Its lexical network architecture is illustrated in \fref{fig:multilink}.
Crucially, it has been designed and implemented as a computational model from the beginning.
This has allowed us to easily explore model variants by simulating empirical data, as well as to analyse what effects
model extensions have on its goodness-of-fit with those data.
Previous experiments have revealed that Multilink's simulation output correlates highly
with existing empirical data for lexical decision and naming tasks \cite[p.~411]{dijkstra2010towards}.

Multilink traces its roots to the Bilingual Interactive Activation Plus model (BIA+, \citeNP{dijkstra2002}),
which was in turn based on the Interactive Activation model (IA, \citeNP{mcclelland1986parallel}) and
the Bilingual Interactive Activation model (BIA, \citeNP{vanheuven1998}).
Like its predecessors, Multilink bases word selection on orthographic activation.
This is the case in tasks like language-specific and general lexical decision.
However, for other tasks it may also be based on
sufficient semantic activation (e.g. semantic categorisation and semantic priming)
or phonological activation (e.g. word naming and word translation).
By allowing multiple read-out codes, the model is able to account for phenomena
such as priming effects and the cognate facilitation effect.

\section{Word activation}

The lexical network represents words by \emph{nodes} of different types:
orthographic, phonological, and semantic.
Two special kinds of nodes are introduced as well: one input node and one language node for each language in the lexicon.

Word representations are linked through bi-directional \emph{connections},
linking orthographic nodes to phonological nodes and vice versa,
as well as linking both orthographic and phonological nodes to semantic nodes.
These connections allow activation to \emph{propagate} through the network.
Finally, language membership is represented by linking orthographic and phonological nodes to their respective language nodes.
Currently, to which language a word belongs does not affect activation within the network.

In order to get activation flowing in the network, the model requires an \emph{input stimulus}.
To represent the input stimulus, Multilink uses the aforementioned input node.
This input node is always maximally activated.
The input information enters the lexical network via its connections to the orthographic nodes.
The strength with which activation propagates to these nodes co-depends on form similarity of the internal representations to the stimulus.

To determine this activation strength, an index of form similarity is required to
reliably compare the input representation to internal representations.
Multilink activates orthographic words based on orthographic similarity, measured in Levenshtein Distance (LD)
between the input and the orthographic representation.
The LD value is normalised over the length of the word symbols involved:
\[
    score = 1 - \frac{dist(source, destination)}{max(len(source), len(destination))}
\]
Here, `dist' refers to the LD function and `len' to the length of the symbol passed.

Essentially, this measure abstracts from the sublexical (grapheme) level found in the IA and BIA+ models.
In doing so, Multilink is able to store and process words of various lengths.
More importantly, by explicitly \emph{avoiding} the use of a slot-based encoding,
activation is not linked directly to letter positions.
Hence, Multilink is able to account for the simultaneous activation
of (partially) embedded words, such as \texttt{ICE} in \texttt{RICE}, and vice versa.
Similarly, this same principle can also account for letter exchanges, like \texttt{JUGDE} for \texttt{JUDGE}.
Furthermore, this characteristic inherently supports the recognition process for (non-identical) cognates.
By definition, such words are similar in orthography between languages,
but are not necessarily of the same length.

For a detailed description of how this is implemented, cf. \citeNP[pp. 8--9]{dijkstra2018multilink}.

\section{Activation propagation}

Having discussed the way representations are activated, we now turn to how activation propagates through the network.
As described, nodes are interconnected through connections.
These connections may be of an \emph{excitatory} (facilitatory) or \emph{inhibitory} (suppressing) nature.
Each of these connections has two weights; one for both directions of the connection.
The values of these weights depend on the types of the two nodes in question.
For example, a connection between orthography and semantics takes weights of the type $OS_\alpha$ or $SO_\alpha$,
depending on the direction.

\pagebreak

Computationally, the propagation of activation is implemented as a two-step process.
This is done so that the order of processing in the computation of activation propagation does not influence activation.

In the first step, the net input is computed by taking the sum over all nodes connecting to the node in question.
This is done by multiplying the connecting nodes' activation by the respecting connection's weight.
In the second step, all nodes are iterated over once more, now applying the activation function over the computed net input.

\begin{figure}[!ht]
    \centering
    \includegraphics[height=1.75cm]{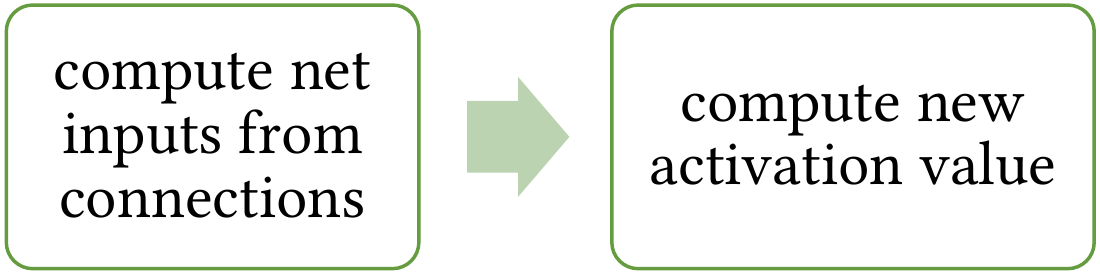}
    \caption{Process diagram for computing node activation.}
    \label{fig:activation_before}
\end{figure}

To illustrate how the activation is propagated through the network,
we will present a simplified account of what happens if we present the stimulus \texttt{AARDE} to the model -- Dutch for `Earth'.
First, the input node symbol (shown at the bottom of \fref{fig:multilink}) is reset to the stimulus.
The input node is connected to all orthographic nodes (\emph{O}) in the network.
To what extent these nodes become active is based on their Levenshtein Distance (LD) with the input node
(cf. \texttt{IO\_alpha} in \fref{app:parameters}).
During each cycle, all nodes whose symbols (partially) match will become slightly more active.
As soon as a node's activation passes the 0.0 point,
it will start to propagate its activation to any connected nodes.
Here, an orthographic node is connected to both its phonetic counterpart (\emph{P}), as well as a semantic concept node (\emph{S}).
This means that, as the orthographic node becomes more active, so will these connected nodes.
In turn, the \emph{S} nodes are connected to not just the Dutch \emph{O} and \emph{P} nodes,
but also their English counterparts.
Hence, as the \emph{S} node for Earth becomes more active, so will \texttt{EARTH} (\emph{O}) and \texttt{3T} (\emph{P}).
Finally, once these nodes have passed the activation threshold, the relevant node is selected by the task/decision system.

This selection mechanism works efficiently for most words.
However, it is unable to correctly predict the translation outcome for interlingual homographs, e.g. \texttt{ROOM}.
\shortcite{goertz2018, dijkstra2018multilink, vanlangendonckIP}.
Consider the situation when the stimulus \texttt{ROOM} is presented to the model for translation.
Both the English word \texttt{ROOM}$_{EN}$ and the Dutch word \texttt{ROOM}$_{NL}$ (meaning cream) will be activated orthographically,
depending on their relative subjective frequency.
Next, the orthographic representations will activate their respective phonological and semantic nodes.
Thus, the concepts \texttt{CREAM}$_S$ and \texttt{ROOM}$_S$ are both activated.
In turn, these concepts will both activate the orthographic and phonological nodes they are linked to.
For this bilingual model, there are \emph{two} phonological nodes per concept.
Hence, in this instance, there will be \emph{four} phonological nodes competing for selection!
The model currently lacks a criterion-based selection mechanism to choose the correct translation in such situations.
Even if we instruct the model to select an output node from a pool belonging to the other language,
a phonological node representing the false friend may be selected instead of the correct translation.
We will return to this problem in chapter \ref{ch:word_translation}.

\pagebreak
\section{Bilingual lexicon}

Like BIA+, Multilink uses an integrated bilingual lexicon for its lexical network, provided in CSV format.
This lexicon currently consists of 1,540 word pairs, whose word length varies between 3 and 12 characters.
These stimuli combine the Dutch Lexicon Project (DLP, \shortciteNP{keuleers2010dutch})
and English Lexicon Project (ELP, \shortciteNP{balota2007english}),
both of which provide behavioural data (reaction times) for all stimuli.
All orthographic readings are complemented with phonetic readings in SAMPA notation,
obtained from the CELEX database \shortcite{baayen1995celex}.

To account for frequency effects, word occurrences per million are included.
These were obtained from the SUBTLEX databases \shortcite{keuleers2010subtlex, keuleers2012british}.
To simulate unbalanced bilinguals, frequencies for English are currently divided by four.
For a detailed account of how these lead to Resting-Level Activations (RLAs), cf. \citeNP[pp. 7--8]{dijkstra2018multilink}.

The first ten rows of the lexicon are printed in \fref{tab:lexicon}.

\begin{table}[!ht]
    \centering
    \begin{tabular}{lrlrlrlr}
        \toprule
        Dutch:O   &         & Dutch:P   &         & English:O  &       & English:P &       \\
        \midrule
        AANBOD    & 26.85   & ambOt     & 26.85   & OFFER      & 18.68 & Qf@R      & 18.67 \\
        AANDACHT  & 56.69   & andAxt    & 56.69   & ATTENTION  & 24.67 & @tEnSH    & 24.67 \\
        AANDEEL   & 9.95    & andel     & 9.95    & SHARE      & 17.38 & S8R       & 17.38 \\
        AANLEG    & 2.88    & anlEx     & 2.88    & INSTANCE   & 4.20  & Inst@ns   & 4.20  \\
        AAP       & 28.56   & ap        & 28.56   & MONKEY     & 8.38  & mVNkI     & 8.38  \\
        AARD      & 15.32   & art       & 15.32   & NATURE     & 11.29 & n1J@R     & 11.29 \\
        AARDAPPEL & 3.34    & ardAp@l   & 3.34    & POTATO     & 2.82  & p@t1t5    & 2.82  \\
        AARDBEI   & 1.56    & ardbK     & 1.56    & STRAWBERRY & 1.38  & str\$b@rI & 1.38  \\
        AARDE     & 100.07  & ard@      & 100.07  & EARTH      & 24.87 & 3T        & 24.87 \\
        AARDIG    & 191.95  & ard@x     & 191.95  & FRIENDLY   & 6.51  & frEndlI   & 6.51  \\
        \bottomrule
    \end{tabular}
    \caption{The first ten rows of Multilink's Dutch-English bilingual lexicon.
        Word frequencies are occurrences per million; orthographic and phonetic representations use the same frequencies.
        English frequencies have been artificially lowered per construction.}
    \label{tab:lexicon}
\end{table}

\section{Task/decision system}

The lexical network is one of the principal components of the Multilink model. 
However, this network alone is not enough to produce output.
This task is delegated to Multilink's \emph{task/decision system} \cite[p.~10]{dijkstra2018multilink}.
The effects of various experimental settings can be investigated in different simulations.
Specifically, participants are tasked with producing different kinds of output based on these settings.
To simulate this process, the task/decision system considers different nodes based on the task at hand.
Similarly, the model's output (but not its network activation) changes based on the task in question.

\begin{figure}[!t]
    \centering
    \includegraphics[width=0.9\linewidth]{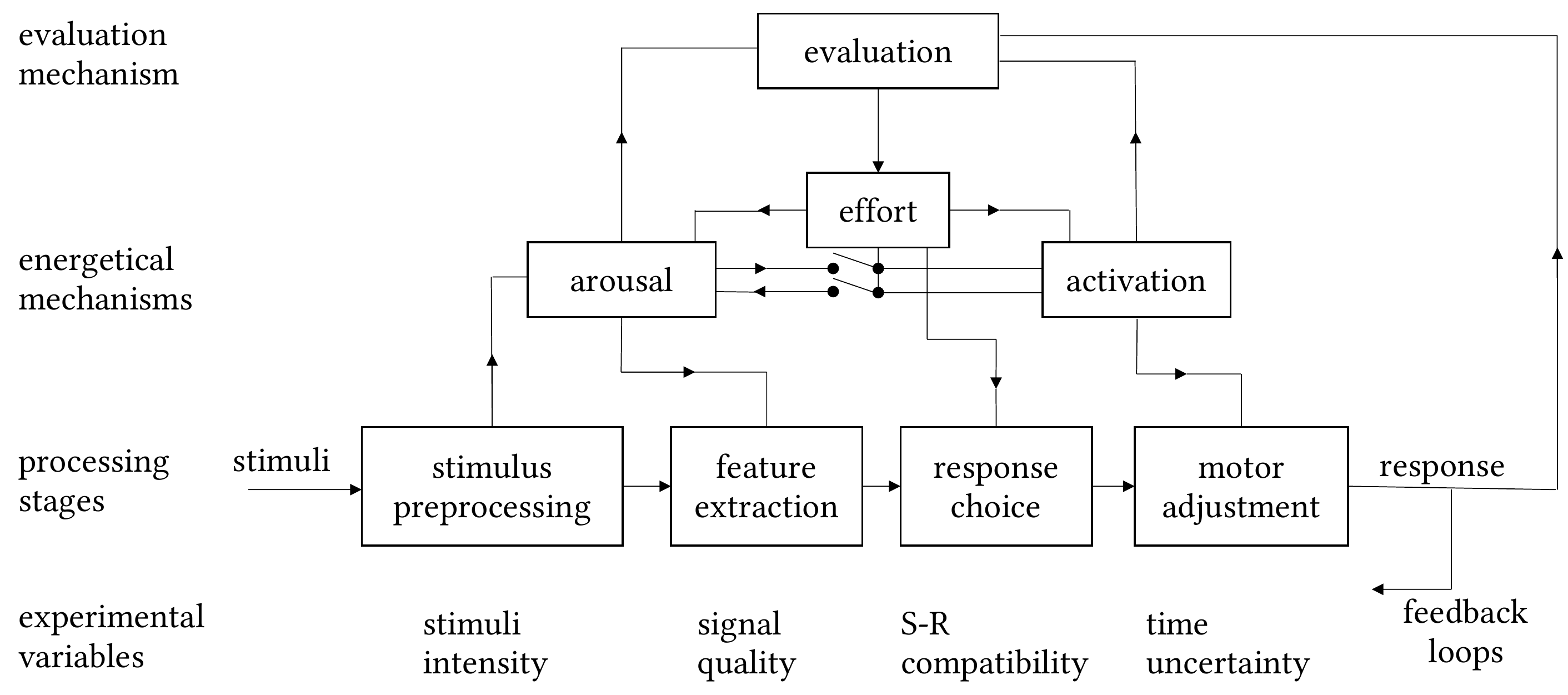}
    \caption{The linear stage model of human information processing and stress, as put forward by \protect\citeA{sanders1983towards}.
        It incorporates linear processing, as well as a parallel energetic and evaluation mechanism.}
    \label{fig:task_sanders}
\end{figure}

To illustrate, consider a lexical decision experiment.
Generally, a participant's only output is a \texttt{YES} or \texttt{NO} response
indicated by a press on one of two associated buttons.
To simulate this, all orthographic nodes in the network are considered to determine the output response.
Once a particular node reaches the critical activation threshold within the cycle time limit,
a \texttt{YES} response is returned.
If the critical threshold is not surpassed within the allotted time limit, a \texttt{NO} response ensues.

In contrast, a naming task requires the same participant to retrieve phonetic and phonological information.
To simulate this retrieval process, the network propagates activation from orthographic nodes to semantic nodes,
which in turn activate the phonological nodes.
These phonological nodes are then considered for output by the task/decision system.
Once a particular node reaches the critical threshold within the cycle time limit,
the corresponding phonological \emph{symbol} is returned as a response.
If not, a \texttt{None} response is returned.

The idea for a task/decision system is not a novel one.
Multilink's is based on BIA+, which in turn borrows ideas from \citeA{sanders1983towards}.
We have reproduced these ideas in \fref{fig:task_sanders} above.
In this figure, the task/decision system is depicted as an evaluation mechanism that regulates arousal,
attentional effort, and activation with respect to the different processing stages of the task at hand.
In this sense, the figure incorporates notions similar to the task schema proposed by \citeA{green1998mental}.
Another idea expressed in this figure is that certain processing stages must necessarily be sequential.
For instance, a motor response can only be given after a response is chosen,
and a response can only be chosen after a set of possible lexical candidates is activated.

In the Multilink model, the lexical network propagates activation regardless of the task at hand.
After enough activation has been propagated through the network, a decision is made based on the task requirements.
This is in line with the notion from \citeA{sanders1983towards} that such mechanisms work in parallel.

    \chapter{Implementing Lateral Inhibition}

All models for word recognition that are presently available in the field of psycholinguistics assume that
when a word is presented, a whole set of lexical possibilities is initially activated.
For instance, hearing the spoken word \texttt{/captain/} results in the activation of all word representations
in the lexicon starting with the onset \texttt{/k/} (like \texttt{CAPTAIN} and \texttt{CAPITAL}),
and reading the printed word \texttt{CORK} activates all words that are orthographically similar
(like \texttt{WORK}, \texttt{COOK}, and \texttt{CORN}).
The general term for such a set is \emph{competitor set}.
In the visual modality, it is often referred to as a \emph{lexical neighbourhood},
while in the auditory modality it is called a \emph{cohort}.
It has often been proposed that these lexical possibilities \emph{compete} for recognition,
i.e.  word form candidates that have been activated on the basis of the input, all affect and \emph{inhibit} each other's activation.
This mechanism is known as \emph{lexical competition} or \emph{lateral inhibition} \shortcite{mcclelland1986parallel, bard1991competition}.
Lateral inhibition leads to a more efficient word recognition process, because by suppressing alternatives, the most active word candidate
(presumably the input word) can be recognised more quickly.

Thus, ideally, introducing lateral inhibition to simulations eases the word selection process:
When more active words inhibit less active words, this theoretically produces one convincing winner more quickly.
Originally, lateral inhibition was not incorporated as a mechanism in Multilink.
When Multilink was first implemented, the decision was made to start with a relatively simple model without lateral inhibition.
This model would then be extended over time \shortcite{dijkstra2010towards, dijkstra2018multilink}.
Surprisingly, Multilink already produced impressive results without lateral inhibition (see \citeNP{dijkstra2018multilink}).
Nevertheless, arguing that empirical studies unequivocally demonstrate the presence of lateral inhibition,
colleagues have criticised the lack of any lateral inhibition in the present version of the model.
In order to incorporate lateral inhibition as a mechanism in the model, Multilink's
lexical network needs to be extended with extra supporting connections between nodes.
This chapter details how this was accomplished, which problems arose as a result, and how they were solved.

The following sections detail how we added an efficient mechanism for lateral inhibition to the \emph{Java} implementation of Multilink.
Benchmarks of the intermediate steps follow in \fref{sec:li_benchmarks}.

\section{Initial implementation}

\begin{figure}[!ht]
    \centering
    \includegraphics[width=0.65\textwidth]{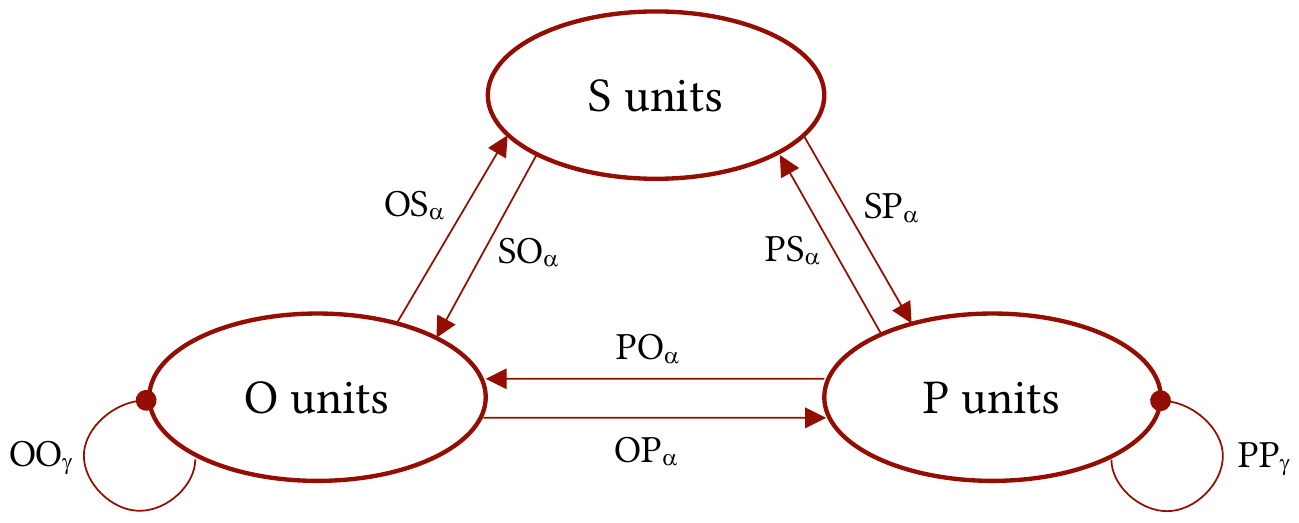}
    \caption{Diagram illustrating the connections between orthographic, phonological, and semantic nodes.
        As language nodes and their connections do not influence node activation at present, they have been omitted for clarity.}
    \label{fig:li_connections}
\end{figure}

To represent lateral inhibition, we introduced two new connection types to the model: OO and PP connections.
In our initial implementation of lateral inhibition, we extended the network by structurally connecting all orthographic nodes with all other
orthographic nodes by means of an OO connection. This was done regardless of the language represented by the node.
We did the same for all phonological nodes, connecting them to all other phonological nodes by means of PP connections.
\Fref{fig:li_connections} illustrates the new model variant.

\pagebreak

\begin{figure}[!b]
    \centering
    \begin{subfigure}[b]{0.49\textwidth}
        \includegraphics[width=\textwidth]{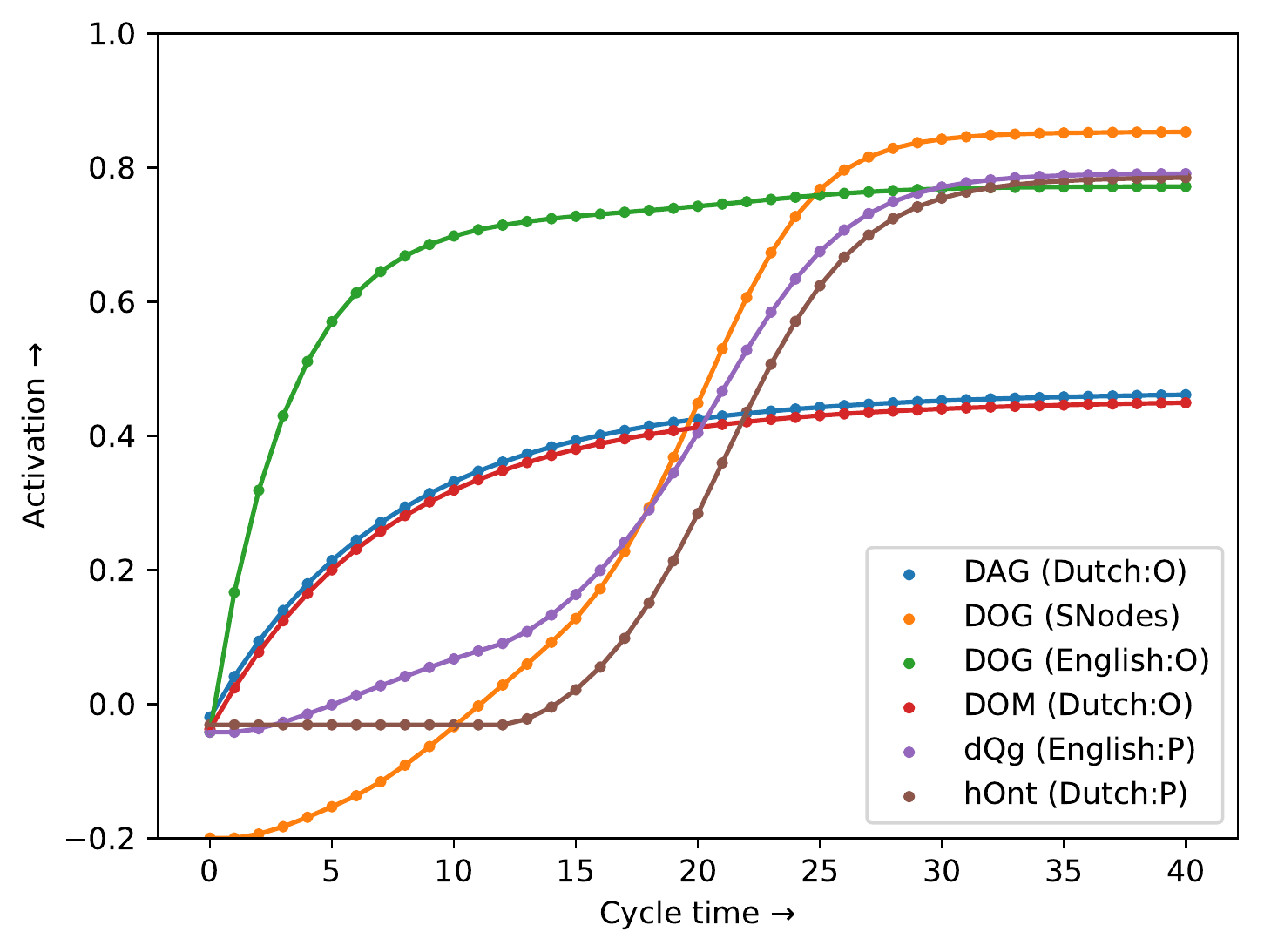}
        \caption{Activation plot for $OO\gamma = 0.0$}
        \label{fig:li_matrix_DOG_0.0}
    \end{subfigure}
    ~
    \begin{subfigure}[b]{0.49\textwidth}
        \includegraphics[width=\textwidth]{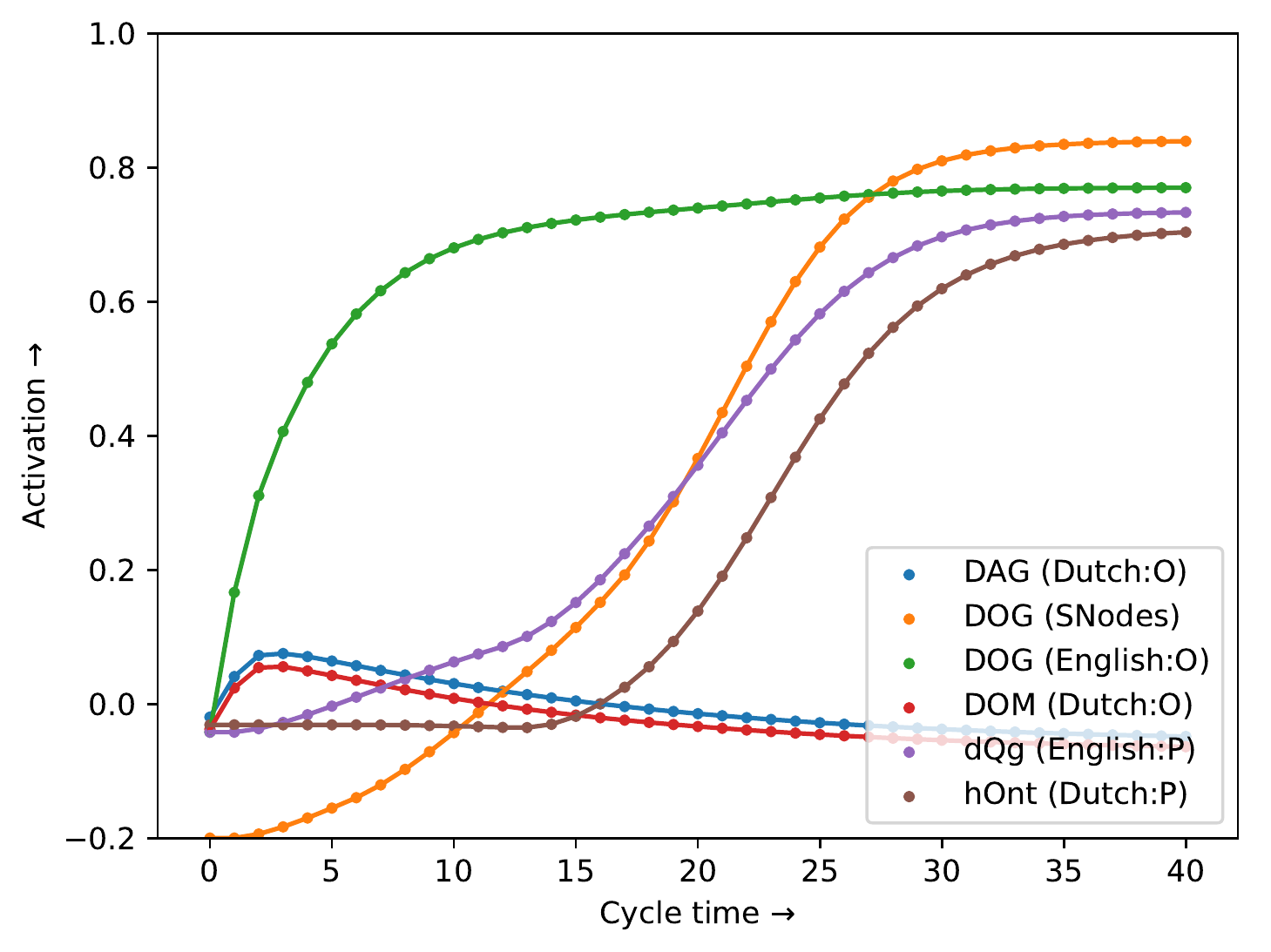}
        \caption{Activation plot for $OO\gamma = -0.1$}
        \label{fig:li_matrix_DOG_0.1}
    \end{subfigure}
    \caption{Chart showing the activation over time for the six most active nodes, given the stimulus \texttt{DOG}.
        Once the semantics for `dog' become active, the phonological node for the relevant Dutch translation
        becomes active as well.
        We observe that adding lateral inhibition (\ref{fig:li_matrix_DOG_0.1})
        leads to suppression of irrelevant neighbours over time.}
    \label{fig:inhibition_dog}
\end{figure}

Initial exploratory simulation on the inhibitory effects in the model looked very promising.
As an example, consider a side-by-side comparison for the stimulus \texttt{DOG} in \fref{fig:inhibition_dog}.
As shown in the chart representing a simulation without lateral inhibition,
both our target stimulus and its neighbourhood competitor words became active over time.
While the more relevant node would ultimately be selected, its competitors were not inhibited (\ref{fig:li_matrix_DOG_0.0}).
This changed when inhibition was introduced:
in the second diagram,
there is a clear effect of inhibition exerted by the target word on the activation of the two other orthographic nodes,
\texttt{DAG} and \texttt{DOM} (\ref{fig:li_matrix_DOG_0.1}).
Note that both of these nodes differ with \texttt{DOG} in only one character -- they are neighbours.
Hence, they are co-activated.

Clearly, this naive approach to implementing lateral inhibition was functioning well.
It also provided us with a relatively straightforward explanation of what activation changes might be happening in the mental lexicon.
However, as we will see in the next section, it also had a rather unpleasant downside.

\section{Connection complexity}

Most of the connectivity in the model is \emph{sparse}.
For example, every orthographic node is connected to only one phonetic node (OP), one language node (OL), and one semantic node (OS).
The opposite is true for inhibitory connections: all orthographic nodes are connected to all other orthographic nodes.
For a summary of the number of outgoing connections per node, please see \fref{tab:li_num_connections}.

This \emph{dense} connectivity introduces a problem in computational complexity:
with lateral inhibitory connections, the number of connections is no longer linear in the number of nodes.
Instead, their number now grows \emph{exponentially} with the length of the lexicon.

\begin{table}[!ht]
    \centering
    \setlength\tabcolsep{4.5pt} 
    \begin{tabular}{lrrrrrrrr}
        \toprule
                      &            & \multicolumn{7}{c}{Number of outgoing connections per node} \\
        \cmidrule(r){3-9}
        Node type     &   \# Nodes & OP/PO & OS/SO & SP/PS & LO/OL & LP/PL &    OO &    PP \\
        \midrule
        Orthographic  &      3,000 &     1 &     1 &     0 &     1 &     0 & 2,999 &     0 \\
        Phonetic      &      3,000 &     1 &     0 &     1 &     0 &     1 &     0 & 2,999 \\
        Semantic      &      1,500 &     0 &     2 &     2 &     0 &     0 &     0 &     0 \\
        Linguistic    &          2 &     0 &     0 &     0 & 1,500 & 1,500 &     0 &     0 \\
        \bottomrule
    \end{tabular}
    \caption{Number of connections per node type by connection type. Note the inhibitory connections on the right-hand side (OO, PP).}
    \label{tab:li_num_connections}
    \setlength\tabcolsep{6pt} 
\end{table}

This is a severe problem. To illustrate its gravity, consider the pool of orthographic nodes.
For a lexicon with 1,500 word pairs, this pool will consist of 3,000 nodes.
To account for the fundamental triangle of $\alpha$--connections (cf. \fref{fig:li_connections}),
we need only 9,000 connections:
\[
3,000 \times ([\textrm{OP/PO}] + [\textrm{OS/SO}] + [\textrm{LO/OL}]) = 3,000 \times (1 + 1 + 1) = 9,000\ \textrm{connections}
\]

\noindent
This amount pales in comparison to the number of OO connections required for lateral inhibition:
\[
3,000 \times [\textrm{OO}] = 3,000 \times 2,999 = 8,997,000\ \textrm{connections}
\]

\noindent
And these are only the connections for orthographic inhibition!
The same number of connections is required to facilitate phonological inhibition as well.
However, all connections we describe here are \emph{mono-directional}.
This means that, like arrows, they point from one node to another, but not necessarily the other way around.
Note that in our \emph{Java} implementation of the model, we implement them as \emph{bi-directional} connections.
This makes the connections symmetrical, but does not necessarily give them same weight in either direction.
Importantly, however, this \emph{reduces} the \emph{spatial complexity} by half.

Nevertheless, even with bi-directional connections, we still have millions of connections to work with.
Considering every connection is checked during \emph{every} network iteration (time cycle),
the introduction of lateral inhibition clearly presents an unworkable regression.
How can this situation be improved without losing the model's inhibitory properties?

\pagebreak

\section{Heuristics}

Inspecting the lexical landscape, we observed that many words in our lexicon are never co-activated.
There are two reasons for this: lack of word form overlap (orthographic or phonological), and lack of meaning overlap (semantic).

Inherent to the activation function used to stimulate nodes in the orthographic pool,
words with no orthographic overlap will not co-activate together, e.g., \texttt{DOG} -- \texttt{PIG}.
This is due to the Levenshtein distance measure used:
words for which the input would effectively need to be entirely rewritten
will not be activated by this measure.

However, we may still find co-activation of such words despite there being little to no overlap.
This is the case when activation is propagated through the semantic network.
For instance, the pair \texttt{BAND} -- \texttt{TIRE} has no orthographic overlap,
but the two items will co-activate due to their semantic equivalence.

Reasoning that nodes that do not co-occur should not influence each other's activation processes,
we decided to use these properties as heuristics to reduce the number of connections.
Before adding a connection, we applied the activation function to the concerning word pair.
If the resulting value was less than a predefined weight constant,
\emph{and} there was no semantic path between the two nodes,
we skipped creating the connection entirely.

Initial findings suggest a weight of 0.001 leaves out enough inhibitory connections for the results to be nearly unaffected,
while a weight of 0.0001 leaves out more connections at the expense of obtaining only near-identical results.
\Fref{tab:li_heuristics} shows the number of bi-directional connections left after applying these heuristics.
Impressively, we can leave out between 26\% and 53\% of the connections to obtain results nearly identical to the baseline.

Unfortunately, connections in the order of millions remain and, as a result, the process of computing activations is still very slow.
We need a better solution.

\begin{table}[!ht]
    \centering
    \begin{tabular}{lrr}
        \toprule
        Connection type            & Cardinality & \\
        \midrule

        \emph{Baseline} \\
        \hspace{1em}OO connections & 4,295,380   & (1,466 $\times$ 1,465 $\times$ 2) \\
        \hspace{1em}PP connections & 4,295,380   & (1,466 $\times$ 1,465 $\times$ 2) \\

        \emph{Weight 0.001} \\
        \hspace{1em}OO connections & 1,138,027   & (avg. 776) \\
        \hspace{1em}PP connections & 790,351     & (avg. 539) \\

        \emph{Weight 0.0001} \\
        \hspace{1em}OO connections & 2,313,698   & (avg. 1,578) \\
        \hspace{1em}PP connections & 1,342,901   & (avg. 916) \\
        \bottomrule
    \end{tabular}
    \caption{Number of bi-directional connections after applying co-activation heuristics.}
    \label{tab:li_heuristics}
\end{table}

\section{Data structure efficiency}
\label{sec:hashmaps}

In order to compute node activation, the Multilink model first computes \emph{input} from \emph{incoming connections}.
It is at this step where the number of connections is most detrimental to model performance.
Even after applying heuristics, the model has to process millions of connections for every iteration.
Not all of these connections are relevant --
only connections to nodes whose activation \emph{exceeds} the 0.0 mark actually influence the target node.
However, currently, the model has no way of knowing which of the connections are relevant.
As a result, to find these connections, the model has to iterate over the \emph{entire} list,
checking the node activation for each connection involved.
What if we could \emph{only} consider the connections for \emph{active} nodes?

Multilink's \emph{Java} implementation assigns ownership of connections to the nodes involved.
Before, this meant nodes had a list containing the few connections it was assigned to.
Now, this list contains thousands of connections, most of which are irrelevant.
\emph{Selecting} the relevant connections for one node is therefore linear at best.
However, doing this for all nodes quickly scales to a quadratic process at least:
given $V$ nodes, $E$ connections, and $T$ cycle times, the model needs $V \times E \times T$ iterations
to compute activation over time for a particular stimulus.
If we can change this selection process to be more efficient, we would solve the speed problem.

A crucial property of the lexical network is that for every node in the network,
this node has \emph{at most} one connection with every other node.
In other words, a node \emph{cannot} have two or more connections with the same target node.
For example, all orthographic nodes are only connected with each other in an inhibitory fashion,
and no other kind of connection.
Concretely, this property means that we can \emph{change} the connection list to a \emph{more efficient data type}:
the \emph{hash map}.

\subsection{Hash maps}

Hash maps (cf. \shortciteNP[pp. 256--260]{cormen2009}) use a \emph{hash function} to map one kind of data object onto another.
In the case of our lexical network, this implies we can know in \emph{constant time} whether or not
a node has a connection with a particular other node.
This, then, allows us to compare a node's connection list to a list of nodes active in the network.

There is one caveat, however: node objects can be \emph{quite complex} and therefore take time to hash.
During our earlier investigations, we observed that the standard hash functions introduced unexpected computational overhead.
To alleviate this problem, we assigned a unique, sequential integer to every node at model creation.
This integer is then used to identify nodes instead, \emph{simplifying} the hashing process considerably,
and thereby solving the hashing problem.

The final ingredient of our solution, then, is to keep a list of active nodes within the network.
We have implemented this list in algorithmically constant time as well.
By comparing a node's current activation to its previous activation, we can easily check whether it went from inactive ($\leq 0.0$) to active.
If that is the case, we add it to the list.
If the opposite is the case, we can assume that it was previously added to the list, and simply remove it.
If its status has not changed, we do nothing.
Hence, we manage a list of active nodes that we can now pass to the input computation function.

Implementing these changes, we found the model's runtime performance to be faster than it had ever been.
However, this was a Pyrrhic victory, because the cost of building the hash maps was quite high:
it took about 4 minutes to \emph{build} the model, rather than the 10 seconds it took before.
We will show this in more detail in \fref{sec:li_benchmarks}.

Nevertheless, using hash maps was very promising, and we set out to find a compromise solution that still
leveraged their power without the cost of long model building times.

\section{Algorithmic approach}

In the previous two sections, we have discussed several improvements to the Multilink implementation.
Notably, the model now actively keeps track of nodes active in the network.
Moreover, as a result of changing data structures, these can be used to more efficiently compute node inputs.
These changes led to our final, fundamentally different implementation of lateral inhibition.

As we have alluded to previously, unlike other connection types, the inhibitory connections are \emph{not} sparse, but dense.
In practice, this means \emph{all} nodes of a certain type are connected to all other nodes of the same type.
Crucially, all inhibitory connections share the same \emph{weights} through parameter values; only the origin and target nodes differ.
This stands in stark contrast to other connections.
For instance, the weights for IO$_\alpha$ and SS$_\alpha$ connections depend on orthographic and semantic similarity,
respectively.

\begin{figure}[!ht]
    \centering
    \vspace{1em}
    \includegraphics[height=1.75cm]{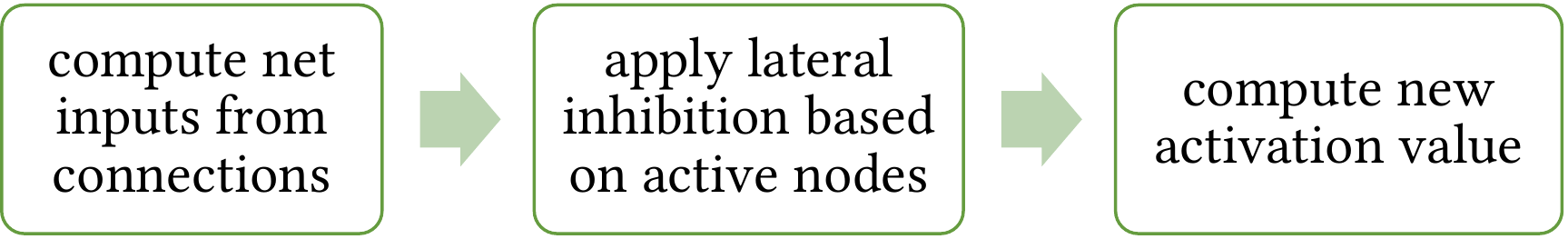}
    \vspace{1em}
    \caption{Updated process diagram for computing node activation.
        The second step is the newly-introduced step dedicated to applying lateral inhibition.
        (Compare with \fref{fig:activation_before}.)}
    \label{fig:activation_after}
\end{figure}

These shared weights, combined with the denseness argument, led to the observation that
we do not \emph{need} connections to achieve lateral inhibition.
Instead, we can apply lateral inhibition for all~\emph{active~nodes} in a separate step in the process of computing node activation.
This new step is set between computing net inputs from connections and computing the new activation value.
\Fref{fig:activation_after} illustrates this.

\subsection{Final solution}

For our final solution, we remove \emph{all} OO$\gamma$ and SS$\gamma$ connections from the network entirely,
including our heuristics module.
Instead, we apply the \emph{effect} these connections would have had \emph{ad hoc},
on top of the net input computed in the first computation step.

Implementation-wise, this means that, when computing the activation for a particular node,
we now simply pass a list of all active nodes and the inhibition parameter for the node type in question.
Each of the applicable nodes will then have its activation applied as inhibition to the node.
The \emph{Java} implementation for this is remarkably short; we have included it in \fref{lst:java_lat_inh}.

\begin{listing}
    \centering
    \singlespacing
\begin{minted}{java}
public void applyLateralInhibition(HashMap<Node, Node> activeNodes, double connWeight)
{
    for (Node other : activeNodes.values())
    {
        // Don't apply lateral inhibition to the same node.
        if (other.equals(this))
            continue;

        // Apply inhibition from nodes of the same type only.
        if (other.getPool().getTypes() != pool.getTypes())
            continue;

        // Proportionally apply the other node's activation as inhibition.
        netInput += other.getCurrentActivation() * connWeight;
    }
}
\end{minted}
    \onehalfspacing
    \caption{Lateral inhibition as implemented in Java.}
    \label{lst:java_lat_inh}
\end{listing}

Importantly, this new implementation shows results that are \emph{identical} to those of the baseline model,
for both the variant with and without lateral inhibition.
Moreover, the network no longer potentially missing any inhibition due to heuristic trickery.
For instance, applying (semantic) priming studies might cause co-occurrences unforeseen by any implemented heuristics.
This, then, alleviates any doubt about future discrepancies in this respect.

Compared to each of the previous approaches, this new approach is surprisingly fast and easy on system memory.
We note that we have kept the hash map discussed in \fref{sec:hashmaps}; the benefits it provides are measureable,
even when the network only contains sparse connections.
Importantly, we find it is as accurate as, yet much faster than, both of the baseline models.
We will discuss this extensively in the following sections.


\section{Benchmarks}
\label{sec:li_benchmarks}

To put our changes to the test, we performed benchmarks on the Centre for Language Studies' computational cluster, Ponyland.
We exclusively used one particular cluster node (\texttt{mlp08}, `featherweight'),
which was not performing any other tasks at the time.
This node uses an Intel\textregistered~Xeon\textregistered~E5-2650 CPU (2.60GHz; 32 threads; 20MB cache)
with 256GB of RAM available.

Four tests were run sequentially under six conditions, each repeated five times.
Where lateral inhibition is used, the parameters $OO_\gamma$ and $PP_\gamma$ are set to $-0.1$.
The average running times of these 24 jobs are included in \fref{tab:li_benchmarks}.

\begin{table}[!hb]
    \centering
    \begin{tabular}{lrrrrr}
        \toprule

        Input                              & \emph{null} & DLP (2) & DLP (10) & Full DLP (1,424) & Stim. avg \\
        \midrule

        \emph{Baseline} \\
        \hspace{1em}Without LI             &         4.0 &    11.4 &     33.6 &     1h 09m 38.4 &     2.931 \\
        \hspace{1em}Initial LI impl.       &        13.8 & 1m 28.2 &  6m 12.0 &    13h 40m 01.0 &    34.541 \\

        \emph{Improvements} \\
        \hspace{1em}LI heuristics          &        21.6 &    52.9 &  2m 42.2 &     5h 23m 45.6 &    13.626 \\
        \hspace{1em}LI heuristics+hashmap  &     4m 15.2 & 4m 17.2 &  4m 19.7 &         8m 52.8 &     0.195 \\

        \emph{Final implementation} \\
        \hspace{1em}Without LI             &         4.4 &    10.7 &     30.4 &        46m 23.8 &     1.951 \\
        \hspace{1em}With LI                &         4.3 &     6.1 &      8.7 &         3m 50.4 &     0.158 \\
        \bottomrule
    \end{tabular}
    \caption{Benchmarks for our implementations of lateral inhibition, measuring how long it takes to process a stimulus list.
        Durations are in seconds unless noted otherwise.
        Time to process a null input file is included to illustrate Java VM startup time,
        as well as the construction of the model and lexical network.
        Stimulus averages were computed over the difference between full DLP input and null input.}
    \label{tab:li_benchmarks}
\end{table}

\section{Conclusions}

Comparing the benchmark results, the final model was found to become considerably faster,
in particular once several stimuli have been processed.
This is a natural side-effect of the way the Java Virtual Machine (JVM) operates.
As time progresses, the JVM identifies critical code-paths and optimises them for the underlying machine's processor
using \emph{just-in-time} compilation (JIT).
To illustrate this aspect, let us compare the jobs with two inputs to those with ten.
Compared to the first two, it takes the latter less time to process eight more stimuli.
From the full DLP simulation, we find an average of 0.158 seconds per stimulus,
compared to 34.541 seconds in the baseline implementation.

On the basis of these benchmarks, we conclude that our final implementation vastly outperforms the baseline implementation.
Previously, we noted that the addition of lateral inhibition generally slows down the simulated selection process,
with words in denser orthographic neighbourhoods suffering more slowdown than other words.
In contrast to our initial implementation, the final results imply a \emph{faster} decision process when lateral inhibition
is enabled.
This has interesting implications for the response-competition process.
As a result of lateral inhibition being implemented, fewer words are present in the competition process,
thereby reducing system load. It may be noted that a similar system of noise reduction by means of
lateral inhibition may be present in the human nervous system (e.g. \shortciteNP{piai2014distinct}).
Interestingly, this empirically observed phenomenon is now also observed to be beneficial in a model like Multilink.

    \chapter{Fitting Lateral Inhibition}
\label{ch:li_fitting}

As we have seen, an efficient implementation of lateral inhibition in the Multilink model was achieved by
using hash map data structures and activation shortlists.
This model extension aims to improve accuracy of predictions from Multilink simulations compared to experimental data.
However, in order to use lateral inhibition properly, it first needs to be \emph{fit}.
This is done by means of hyperparameters, which adjust the strength of excitatory or inhibitory connections in the model.
For Multilink's implementation of lateral inhibition, these are the inhibitory $OO_\gamma$ and $PP_\gamma$ parameters.
This chapter discusses the fitting process of both of these hyper-parameters
by means of a \emph{grid search} algorithm.

First, we will briefly discuss the grid search algorithm used to perform the parameter search.
We then continue by applying this algorithm to reaction time data from three extensive lexical decision studies:
the English Lexicon Project \shortcite{balota2007english}, the British Lexicon Project \shortcite{keuleers2012british},
and the Dutch Lexicon Project \shortcite{keuleers2010dutch}.
Finally, we will apply the optimal hyper-parameter values we find to simulate results
from a lexical decision experiment focusing on dense neighbourhoods \shortcite{mulder2018}.
As we will see, we find that correlations improve with the introduction of lateral inhibition to the network.

\section{Grid Search}
\label{sec:gridsearch}

We have introduced two parameters for the lateral inhibition process: $OO_\gamma$ and $PP_\gamma$.
However, the question of what values these parameters should take has so far been left unanswered.
Finding these values is important, as they ultimately determine accuracy with respect to simulating experimental data.
To answer this question, we introduce use a grid search algorithm to iteratively explore the values in the parameter domain.
It was decided to perform a fit on empirical data, constraining $OO_\gamma = PP_\gamma$.
Both of these parameters serve separate pools of nodes in the network.
Notably, both pools are of an equal size. By fitting the parameters between O and P symmetrically,
the search space involved is reduced considerably.

\begin{figure}[!ht]
    \centering
    \includegraphics[height=4.5cm]{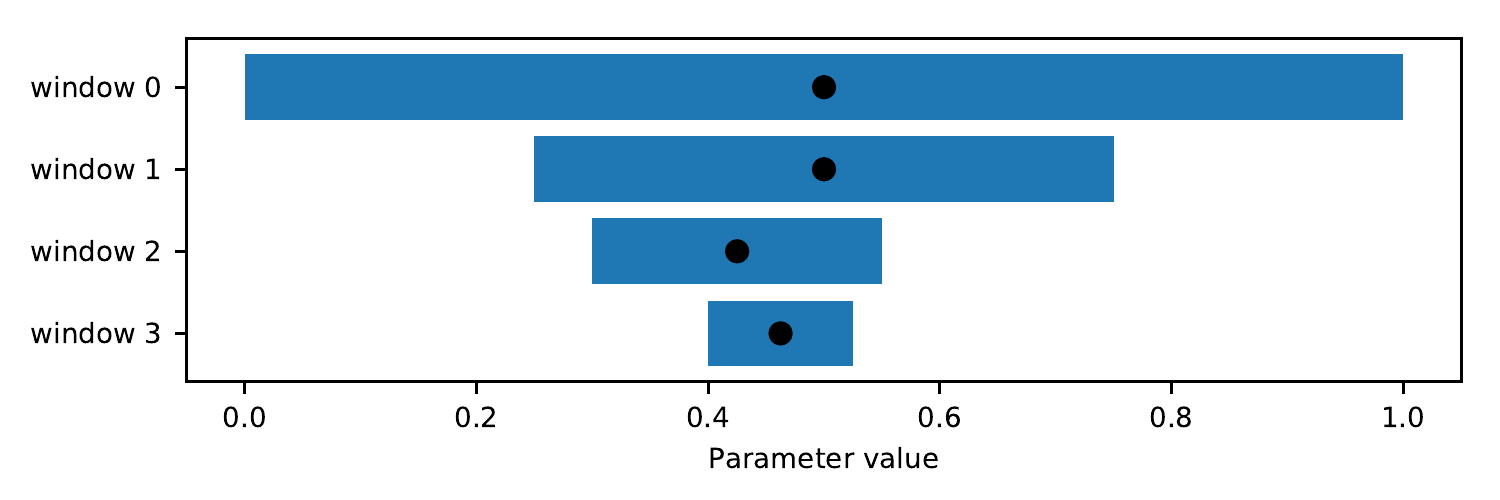}
    \caption{Hypothetical example of a sliding window as used by the grid search algorithm.
        The optimal parameter values encountered by the algorithm are indicated in each window.}
    \label{fig:sliding_window}
\end{figure}

The grid search algorithm applies an iterative \emph{breadth-first} search to the parameter domain.
This search is constrained to an iteratively-narrowing window,
with each iteration sampling $N$ \emph{equidistant} points.
Each point is then used as a parameter value in a simulation,
after which the simulation results are evaluated using a \emph{fitness function}.
When the iteration concludes, the optimal fitness value is determined out of the $N$ points considered.
The window is then \emph{halved} in size and \emph{centred} around this optimal fitness point,
after which a new iteration starts.
If the next iteration does not yield an optimal value bigger than $\epsilon$,
the algorithm terminates.
\Fref{fig:sliding_window} illustrates this search process with an example.

The parameter domain ranges from 0.0 (no inhibition) to -1.0 (full inhibition).
Using $N = 20$, this implies an initial step size of 0.05.
Halving the window size for the next iteration means the subsequent step size will be 0.025, et cetera.
As we will see in the next section, we find this value of $N$ provides us with enough data points
to gain insight into the inhibitory mechanisms between nodes in the orthographic pool
and nodes in the phonological pool.

Ultimately, we aim for our algorithm to find parameter values that will see the model yield
patterns similar to experimental behavioural data.
Assuming our model can indeed provide a good fit for these data,
this goal is attainable by structurally evaluating \emph{model-to-data} fitness.
We therefore opted to use the Pearson correlation coefficient as the grid search fitness function,
optimising on positive linear correlations.

A listing of the algorithm as implemented in Python is included in \fref{app:gridsearch}.

\section{Exploratory Results}

We applied the grid search algorithm as described to stimuli and reaction time data from three lexical decision studies
\shortcite{balota2007english, keuleers2012british, keuleers2010dutch}.
The results for simulations using bilingual lexicons are plotted in \fref{fig:fit_bilis} below.
Similar patterns are observed when monolingual lexicons are used.

\begin{figure}[!ht]
    \centering
    {\begin{subfigure}[b]{0.62\textwidth}
        {\includegraphics[height=5cm]{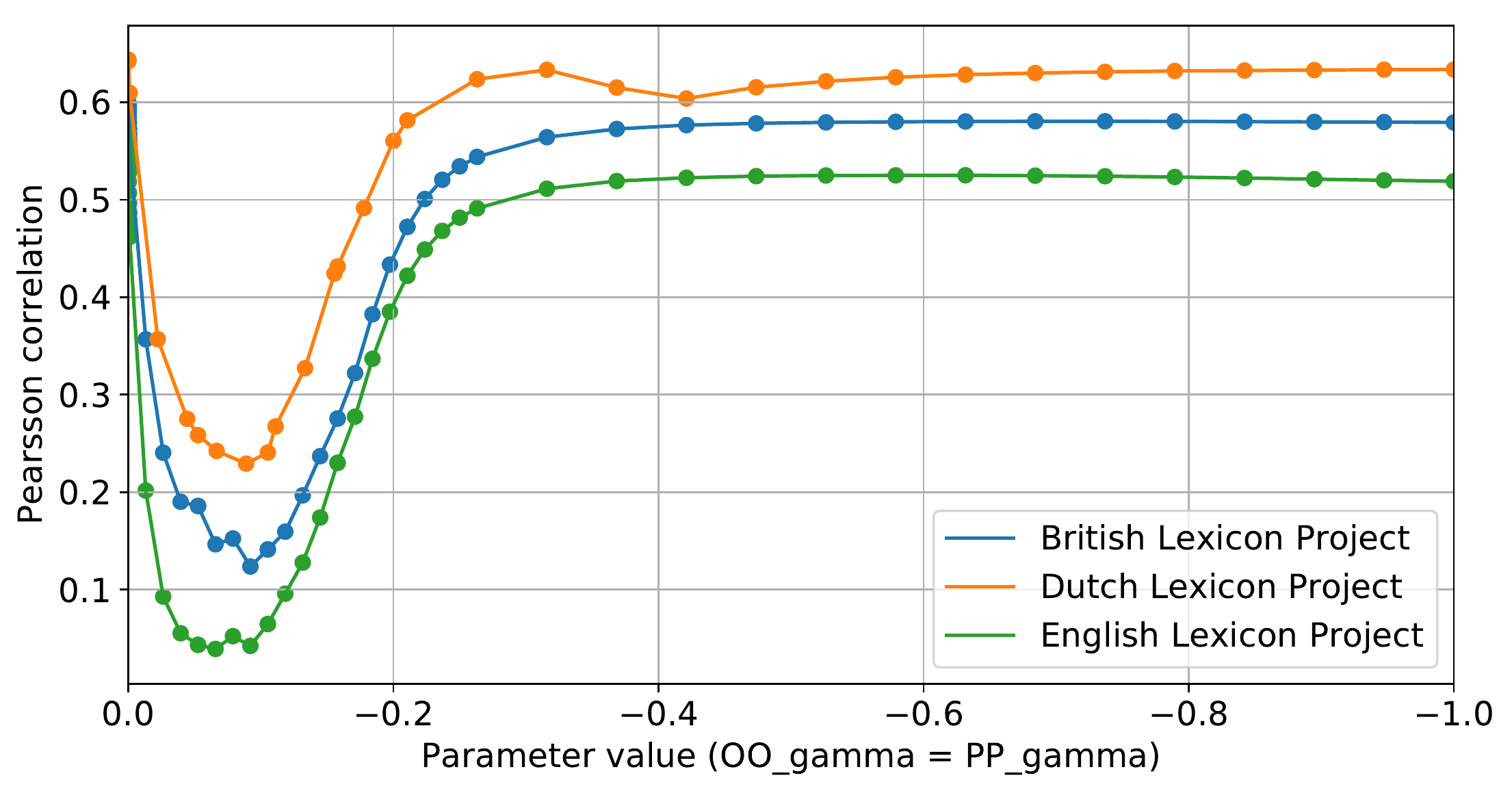}}
        \caption{Grid search results for the whole activation domain.}
        \label{fig:fit_bilis_whole}
    \end{subfigure}}
    ~
    {\begin{subfigure}[b]{0.36\textwidth}
        {\includegraphics[height=5cm]{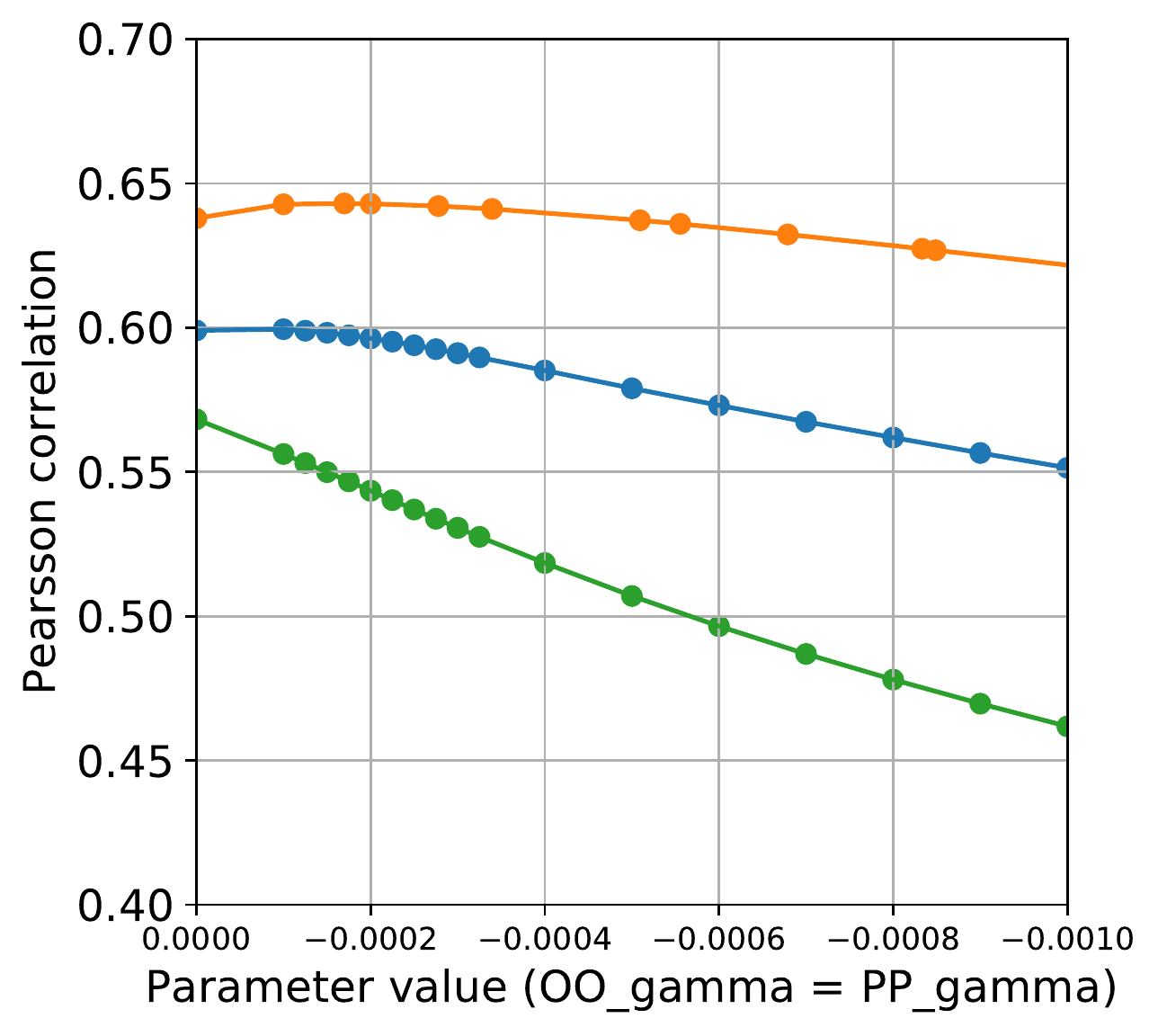}}
        \caption{Zoomed to smaller domain.}
        \label{fig:fit_bilis_zoom}
    \end{subfigure}}
    \caption{Correlations by inhibitory parameter values for simulations using bilingual lexicons, as obtained in the grid search process.
        \Fref{fig:fit_bilis_zoom} zooms in on the `Goldilocks zone' where inhibition appears to be optimal.}
    \label{fig:fit_bilis}
\end{figure}

\subsection{Observing inhibition effects}

\begin{figure}[!ht]
    \centering
    \includegraphics[width=.8\textwidth]{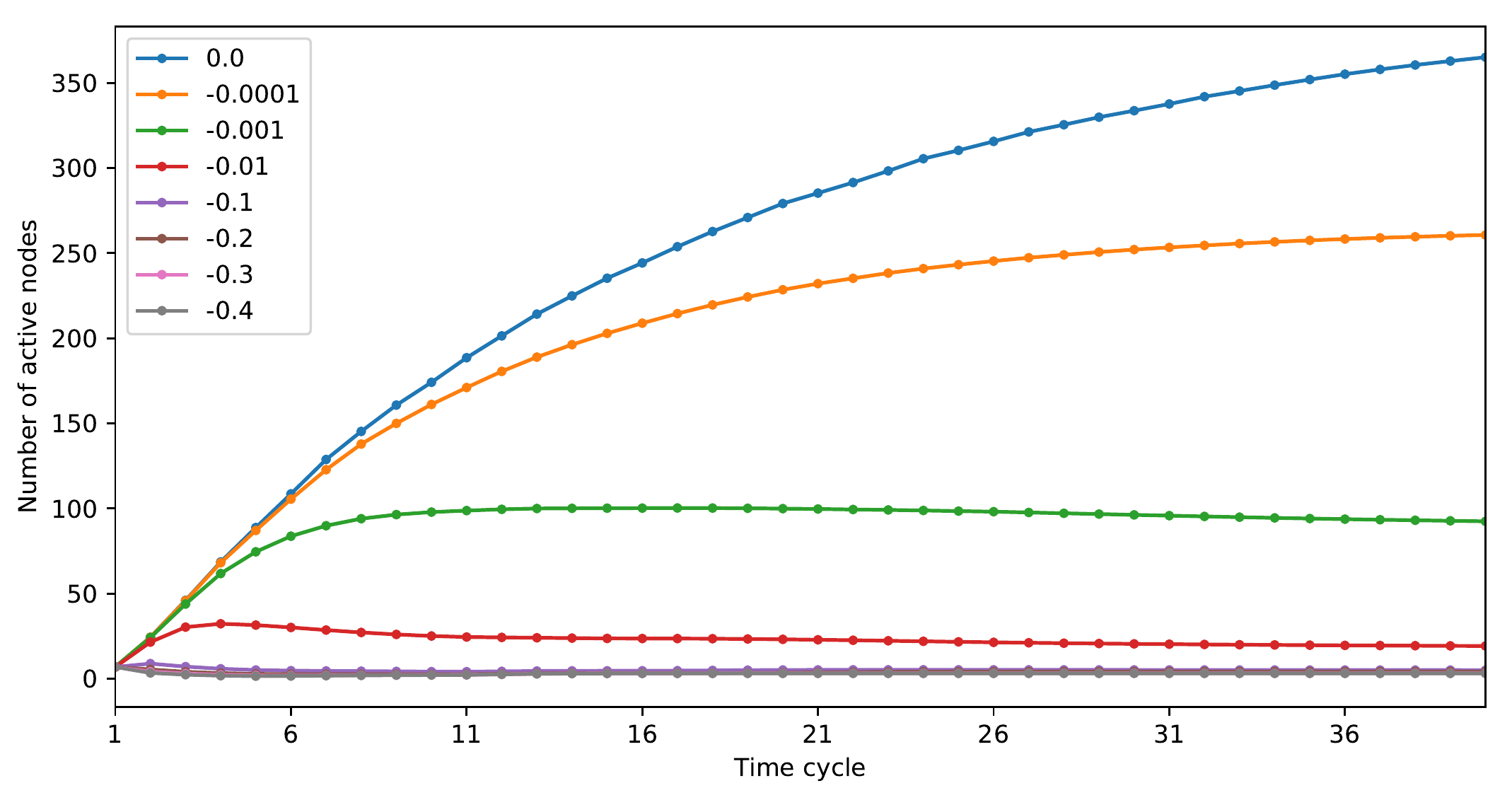}
    \caption{Number of active nodes over time by lateral inhibition setting.}
    \label{fig:fit_li_num_nodes}
\end{figure}

To observe the effect lateral inhibition has on the number of active nodes in the model,
we performed a readouts by node type at the end of the final simulation cycle (\fref{tab:num_nodes}),
as well as over time for all nodes (\fref{fig:fit_li_num_nodes}).

\begin{table}[!ht]
    \centering
    \begin{tabular}{rrrrr}
        \toprule
        LI parameter &   Overall  &  Orthographic  &  Phonological  &  Semantic \\
        \midrule
              0.0    &    365.08  &        311.38  &         51.95  &      1.75 \\
             -0.0001 &    260.64  &        219.22  &         39.71  &      1.71 \\
             -0.001  &     92.43  &         77.27  &         13.63  &      1.53 \\
             -0.01   &     19.21  &         15.62  &          2.41  &      1.18 \\
             -0.1    &      5.03  &          1.92  &          2.07  &      1.04 \\
             -0.2    &      4.18  &          1.15  &          2.00  &      1.03 \\
             -0.3    &      3.24  &          1.13  &          1.08  &      1.03 \\
             -0.4    &      3.09  &          1.04  &          1.03  &      1.02 \\
             -0.5    &      3.03  &          1.03  &          1.02  &      0.97 \\
        \bottomrule
    \end{tabular}
    \caption{Average number of active nodes by lateral inhibition parameter setting, split by type.
        Measurements were obtained at the end of complete network propagation, that is after 40 time cycles.}
    \label{tab:num_nodes}
\end{table}

\section{Analysis}

The results from the grid search simulations show interesting patterns.
All three of the fitted datasets show a similar shape in the first half of the domain:
starting off at a high correlation without inhibition,
we observe correlations decreasing to a valley shape as inhibition is increased.
Beyond the -0.1 mark, correlations rise again.
This last fact is of particular interest to us.
Why do correlations first decrease, before increasing again, as more inhibition is added to the network?

If no lateral inhibition is present in the network (0.0), all neighbouring words remain active relative to their degree of overlap.
Once we introduce a little bit of inhibition (-0.0001), these neighbours start to compete for their activation.
Again, their competing power is relative to their activation.
The effect of this quickly becomes apparent from \fref{tab:num_nodes}:
the number of active nodes quickly drops by a third, even with this little amount of inhibition.
\Fref{fig:fit_li_num_nodes} shows similar effects for the number of all nodes over time.
Moreover, there is a slight, general delay in word recognition speed,
with words in denser neighbourhoods affected to a higher extent.

These competition effects become extreme when the inhibition parameters are increased to -0.1.
As a direct result of the nodes competing at this rate, words in denser neighbourhoods are no longer recognised.
This, then, results in the valley we see in terms of correlations.
Once inhibition increases further still (e.g. to -0.2),
competing neighbours start to become eliminated very early on in the activation process.
Hence, conditions with high inhibition start to resemble situations without any inhibition present (-0.4 and higher).
Adding more inhibition beyond this point hardly seems to matter; the graphs \emph{flatline} beyond this point.

All three datasets share a similar curve with respect to these extreme values.
The Dutch Lexicon Project shows a minor dip around the -0.35 mark, however,
which is absent in the curve for the other two projects.
We observe the same pattern in both monolingual and bilingual versions of the lexicon.
Hence, we speculate this is inherent to the makeup of the Dutch lexicon.
While the lexicon was designed to not be morphologically complex,
a possible explanation is that there are still relatively many Dutch words embedded in other words,
thereby affecting each other.
Further research is required to give a conclusive answer here.

Local inspection suggests the optimal lateral inhibition values to be constrained to the (-0.1, 0.0) interval.
Indeed, we find the highest correlations around the -0.0001 mark (cf. \fref{fig:fit_bilis_zoom}).
As representations begin to compete more, it takes a \emph{stronger input} for them to actually become active.
As a direct consequence, far \emph{fewer} nodes pass the initial activation threshold.
This leads to words not being recognised, or recognised far later than experimental trials show.
Hence, correlations drop rapidly with such parameter settings.

\subsection{Generalisation to other tasks}

How do we interpret this apparent local optimum around the -0.0001 mark, and the flatlining after the -0.4 mark?
Before generalising these findings, we should consider the task demands of the current simulations.
At present, we are simulating a \emph{lexical decision} task.
In Multilink, this task requires that \emph{any} node in a particular orthographic pool passes a particular \emph{threshold}.
As we will see, this is a relatively undemanding task.

\begin{table}[!t]
    \centering
    \begin{subtable}{0.48\textwidth}
        \begin{tabular}{lllll}
            \toprule
            $OO\gamma$ &  all             &  control         &  NC1             &  NC2 \\
            \midrule
            \phantom{-}0.0       &  0.5565          &  0.5425          &  \textbf{0.5218} &  \textbf{0.6265} \\
            -0.0001    &  \textbf{0.5616} &  \textbf{0.5498} &  0.5140          &  0.6247 \\
            -0.001     &  0.4854          &  0.4606          &  0.4537          &  0.5865 \\
            -0.01      &  0.3139          &  0.2824          &  0.3013          &  0.4331 \\
            -0.1       &  0.1277          &  0.1139          &  0.0201          &  0.3094 \\
            \bottomrule
        \end{tabular}
        \caption{English Lexicon Project (ELP) correlations}
        \label{tab:oo_elp_corr}
    \end{subtable}
    \begin{subtable}{0.48\textwidth}
        \begin{tabular}{lllll}
            \toprule
            $OO\gamma$ &  all              &  control          &  NC1              &  NC2    \\
            \midrule
            \phantom{-}0.0       &  0.5818           &  0.5648           &  0.6346           &  0.6382 \\
            -0.0001    &  \textbf{0.5977}  &  \textbf{0.5858}  &  \textbf{0.6390}  &  \textbf{0.6388} \\
            -0.001     &  0.5525           &  0.5377           &  0.6196           &  0.6021 \\
            -0.01      &  0.4401           &  0.4178           &  0.5052           &  0.5020 \\
            -0.1       &  0.2013           &  0.1871           &  0.0896           &  0.3439 \\
            \bottomrule
        \end{tabular}
        \caption{British Lexicon Project (BLP) correlations}
        \label{tab:oo_blp_corr}
    \end{subtable}
    \par\bigskip
    \begin{subtable}{0.48\textwidth}
        \begin{tabular}{lllll}
            \toprule
            $OO\gamma$ &  all              &  control          &  NC1              &  NC2 \\
            \midrule
            \phantom{-}0.0       &  0.6379           &  0.6231           &  \textbf{0.6642}  &  0.7087 \\
            -0.0001    &  \textbf{0.6449}  &  \textbf{0.6327}  &  0.6629           &  \textbf{0.7111} \\
            -0.001     &  0.6165           &  0.6105           &  0.5934           &  0.6896 \\
            -0.01      &  0.4829           &  0.4647           &  0.4616           &  0.5874 \\
            -0.1       &  0.2795           &  0.2642           &  0.3865           &  0.3539 \\
            \bottomrule
        \end{tabular}
        \caption{Dutch Lexicon Project (DLP) correlations}
        \label{tab:oo_dlp_corr}
    \end{subtable}

    \caption{Pearson coefficients between Multilink LeD cycle times and ELP reaction times.
       As $OO\gamma$ and $PP\gamma$ were fit symmetrically,
       $PP\gamma$ of equal strength is implied where $OO\gamma$ is used.
       $OO\gamma = 0.0$ denotes the baseline without any lateral inhibition.}
    \label{tab:oo_elp_blp_dlp_corr}
\end{table}

Concretely, in the ELP simulation, we are looking out for \emph{any} node in the English orthographic pool passing the 0.72 activation mark.
If lateral inhibition is set to higher values, the most activated node quickly inhibits all other nodes of the same type.
Inherent to the activation function Multilink uses, this will be the node with full orthographic overlap.
In a lexical decision task, this implies the node in question will proceed to the 0.72 activation mark,
having effectively eliminated any competing nodes.
Hence, correlations \emph{flatline} beyond the -0.4 point.
Even if correlations are slightly lower there, in essence, they reflect the situation without lateral inhibition present.
This means that we can do very fast approximations of lexical decision studies by using a lateral inhibition value around -0.4.

It is important to note that, in spite of these findings, inherently, these results do not generalise to translation studies.
If lateral inhibition is set to a strong value such that all other nodes of the same type are inhibited,
there is no chance for translation equivalents to become active in the process!
Hence, for general purposes, we aim to find a value that shows inhibitory properties, but not overly so.

As a final question, we consider how the values we have found compare to theoretical considerations.
Recall that Multilink traces its roots back to the Interactive Activation model \shortcite{mcclelland1986parallel}.
Curiously, this model uses a word-to-word inhibition value of -0.21.
Going by our findings, however, this value clearly results in too much inhibition within the Multilink network.
We attribute this to the way word representations are activated within the two networks.
The IA model incorporates sublexical representations combined with a slot encoding,
while Multilink omits these and uses a Levenshtein Distance measure to directly activate orthographic word representations from input.
This difference in a combination of mechanisms may explain the need for a much lower amount of inhibition in Multilink.

\section{Application}

In the empirical literature on word recognition, lateral inhibition is seen as the brain's solution for dealing with competing words.
It speeds up processing and eliminates noise.
The degree of lateral inhibition depends on the number of words that have form overlap with the target word.
When a word has many neighbours, it is located in a dense neighbourhood.
Conversely, words with few neighbours have a sparse neighbourhood.
Extreme cases are hermits: words without any neighbours.

Let us investigate the effects of neighbourhood density in two versions of Multilink,
without and with lateral inhibition, by considering and simulating a recent
lexical decision study that manipulated the neighbourhoods of target words \shortcite{mulder2018}.
The word stimuli in this study were used as input for Multilink with two settings of lateral inhibition:
none at all (0.0) and minimal (-0.0001).
Note that the latter of these was previously found to be optimal in general.
The results of our simulations are presented in \fref{tab:mulder}, both without and with lateral inhibition.
In all but one of the conditions, adding lateral inhibition leads to an improvement in correlations.

\begin{table}[t]
    \centering
    \begin{tabular}{lrrrrr}
        \toprule
                                          &     & Baseline & Minimal          & \multicolumn{2}{c}{Optimal correlation} \\
        \cmidrule(r){5-6}
        Condition                         & $N$ & (LI=0.0) & (LI=-0.0001)     &   LI value & Correlation \\
        \midrule
        Overall                           & 102 & 0.66251  & \textbf{0.67016} &  -0.00017  &  0.67171    \\
        Both Dutch and English Neighbours & 30  & 0.67776  &         0.65399  &   0.00000  &  0.67776    \\
        Complete Hermits                  & 29  & 0.65491  & \textbf{0.66755} &  -0.78532  &  0.69089    \\
        Only Dutch Neighbours             & 14  & 0.78805  & \textbf{0.79267} &  -0.36842  &  0.81210    \\
        Only English Neighbours           & 29  & 0.58240  & \textbf{0.59212} &  -0.26316  &  0.62304    \\
        \bottomrule
    \end{tabular}
    \caption{Results from simulating the second experiment from \protect\shortciteA{mulder2018}.
        Correlations improve for all but one of the individual conditions.
        Note that non-words were left out of the simulations.}
    \label{tab:mulder}
\end{table}

\begin{figure}[!ht]
    \centering
    \includegraphics[width=0.85\textwidth]{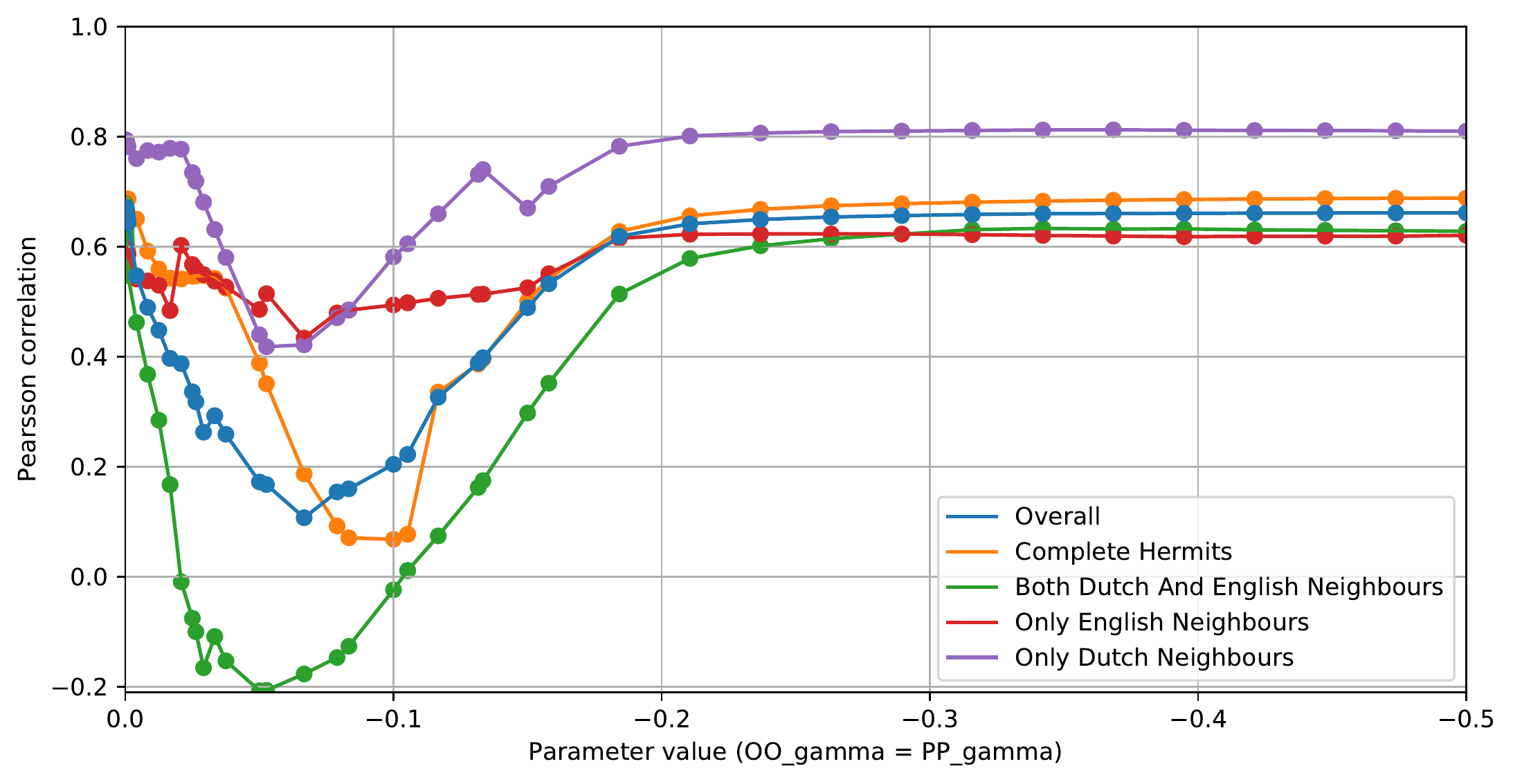}
    \caption{Correlations for Mulder experiment depending on Lateral Inhibition value, split by condition.}
    \label{fig:li_mulder}
\end{figure}

Reassuringly, if we apply the grid search algorithm from section~\ref{sec:gridsearch} to this case,
we find roughly the same optimal value overall.
However, interestingly, for the individual conditions, the optima differ.
These optima are listed in \fref{tab:mulder} as well,
while the full results are plotted in \fref{fig:li_mulder}.

As we hypothesised in the previous section,
we find hermit words, without neighbours, to best withstand lateral inhibition.
Furthermore, we find a different optimum for Dutch and English neighbours.
This to be expected, because we are simulating the performance of late unbalanced bilinguals,
for whom English is a second language.

\section{Conclusions}

Having extended the Multilink model with lateral inhibition, we have now fit the model's accompanying hyper-parameters
to reaction time data from three extensive lexical decision studies:
the English Lexicon Project \shortcite{balota2007english}, the British Lexicon Project \shortcite{keuleers2012british},
and the Dutch Lexicon Project \shortcite{keuleers2010dutch}.
We find an optimal, generalisable parameter set for the lateral inhibition parameters $OO_\gamma = PP_\gamma = -0.0001$.

The number of active nodes in the network steadily increases over time when lateral inhibition is not present.
Conversely, as the amount of inhibition is increased, word form competition becomes stronger.
As a resulting, the number of active nodes decreases.
We find that this substantially reduces the time required to perform simulations.
This is limited to certain tasks, however, as too much competition leads to the inability to perform word translation.
However, for recognition tasks, a parameter value of $OO_\gamma = PP_\gamma = -0.4$ may be used for quick iteration
without adversely affecting correlations.

Finally, we applied the Multilink model with this optimal lateral inhibition setting
to an empirical study involving a lexical decision experiment focusing on dense neighbourhoods \shortcite{mulder2018},
for which we find an overall Pearson correlation coefficient $r = 0.67$.

    \chapter{Word Translation Problems}
\label{ch:word_translation}

Multilink can simulate a variety of experimental tasks, including \emph{lexical decision} and \emph{word naming}.
These two tasks are the most frequently applied experimental techniques in the domain of word recognition.
In a lexical decision task, a participant is presented with a word on a screen and asked to indicate by a press
on one of two buttons whether it exists (`yes') or not (`no') in a particular language (two-alternative forced choice response).
In a naming task, the participants reads out loud as quickly as possible the presented word or letter string.
The two tasks are usually aimed at measuring the speed and accuracy of lexical performance in one particular language.
Both tasks can be performed by both monolingual and multilingual speakers.
However, in the case of multilingual speakers,
words of more than one language may be present in the experiment.
An example of a task that inherently requires handling words of several languages at about the same time is \emph{word translation}.

There are two main experimental variants of the word translation task: word translation \emph{production}
(e.g. \citeNP{degroot1992determinants}) and word translation \emph{recognition} \cite{degroot1995translation}.
In the production variant of the translation task, a participant is presented with one word on a screen,
which can be either from their L1 or an L2.
Participants are tasked with quickly naming the correct translation of this word in the non-presented language.
Alternatively, in the (slower) recognition variant of the translation task, a participant is presented with a pair of words on each trial:
one word from their L1 and one from their L2.
Participants are now asked to decide whether or not these two words are translations of one another.
Rather than indicating this by a spoken response (`yes' or `no'), they can also do this by button press.
Crucially, a consistent finding in these tasks is that cognates are translated faster and more accurately than non-cognates
(e.g. \citeNP{christoffels2006, degroot1994forward}).
Similarly, words that occur with a higher frequency are translated faster than low-frequency words.

Both the lexical decision and word naming tasks have been implemented in Multilink
using a threshold-based \emph{response selection}.
The decision system of each tasks monitors a particular representational \emph{pool} in the lexical network,
and once any node in this pool passes a certain activation value (the threshold),
it is selected as the response.
Concretely, for lexical decision, the orthographic pool of the target language is monitored.
Likewise, for word naming, the phonological pool of the target language is monitored.
In both cases, surpassing an activation threshold of 0.72 is used as a word selection criterion.

This selection mechanism has been found to result in a good match between empirical and simulation results for lexical decision tasks,
word naming tasks, and even word translation tasks \cite{dijkstra2018multilink}.
However, the mechanism was later found to be insufficient to accurately simulate experimental data involving interlingual homographs
(e.g. \shortciteNP{vanlangendonckIP}, \citeNP{goertz2018}).

In this chapter, we will expose the problems of  current model simulations by examples and then discuss our proposed solution.
Finally, we will discuss the efficacy of this solution by applying an implementation thereof
to two datasets.

\section{Interlingual homographs}

Translation problems come to light when the model tries to translate the type of words called interlingual homographs.
Like identical cognates, interlingual homographs are pairs of words that share their full form across languages.
However, unlike cognates, the two readings of an interlingual homograph have a different meaning entirely.
For Dutch and English, examples are \texttt{FILM}, \texttt{ROOM}, \texttt{SLIM}, and \texttt{WET}.

In terms of Multilink's lexical network, interlingual homographs are represented by two orthographic nodes that will receive
roughly equal activation.
In turn, both of these activate their semantics and phonology.
As a consequence, there can be not two, but at least four competing phonological nodes at one moment in time!
For instance, the input homograph \texttt{ROOM} will fully activate the pronunciations
\texttt{/kam@r/}$_{NL}$, \texttt{/krim/}$_{EN}$, \texttt{/rom/}$_{NL}$, and \texttt{/rum/}$_{EN}$!
To simulate performance in word translation tasks, where a participant must pronounce the presented word in the other language,
this is highly problematic: how to decide which of these to utter?
Indeed, in nearly all cases, the earlier version of the model ends up selecting a wrong, competitor candidate.
This could be a candidate of the wrong language, the input word itself,
or perhaps other highly frequent words, like \texttt{ROEM}$_{NL}$.

\begin{figure}[t]
    \centering
    \begin{subfigure}[b]{0.49\textwidth}
        \includegraphics[width=\textwidth]{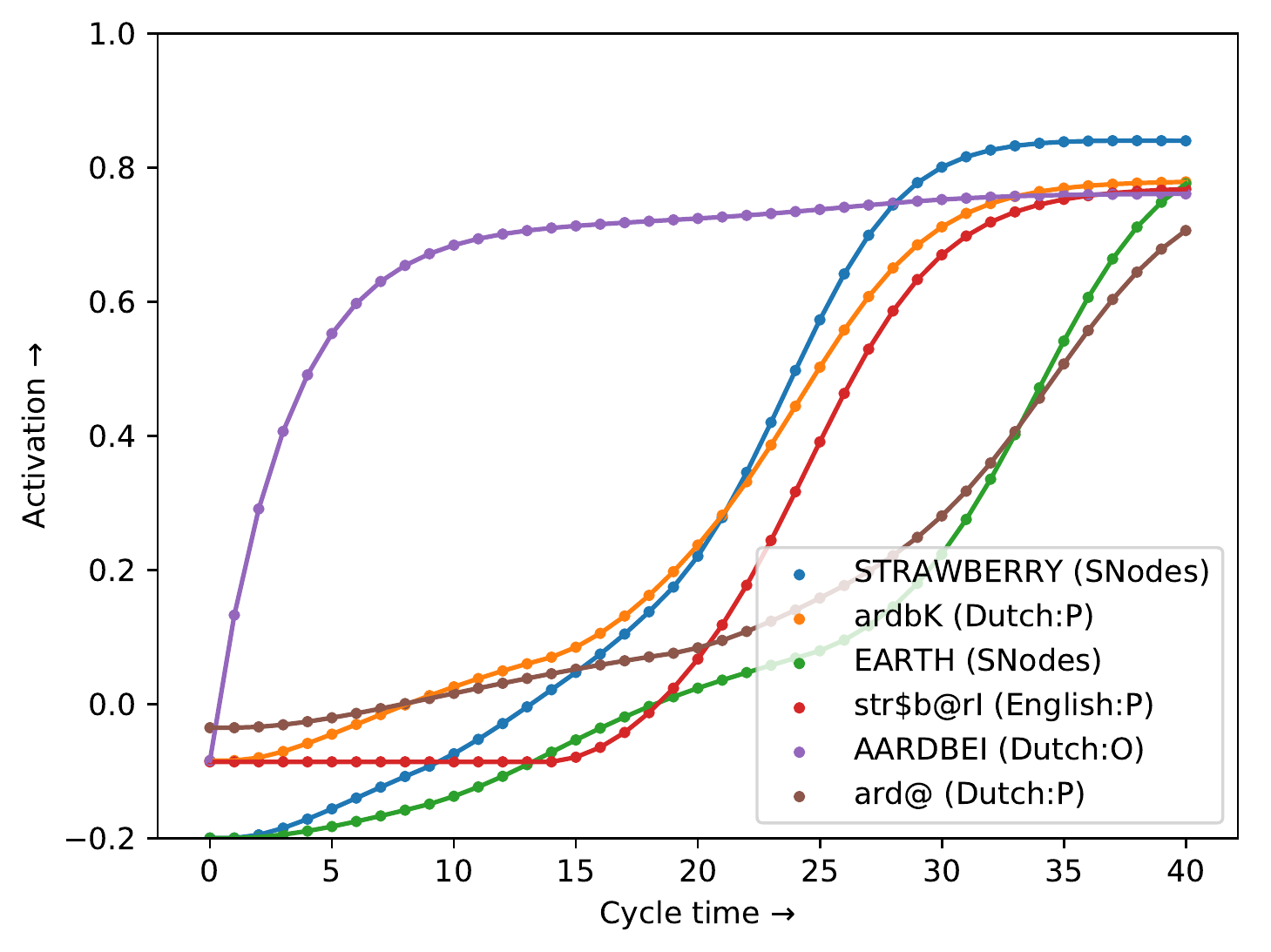}
        \caption{Matrix chart for input \texttt{AARDBEI} (strawberry).}
        \label{fig:wt_matrix_aardbei}
    \end{subfigure}
    ~
    \begin{subfigure}[b]{0.49\textwidth}
        \includegraphics[width=\textwidth]{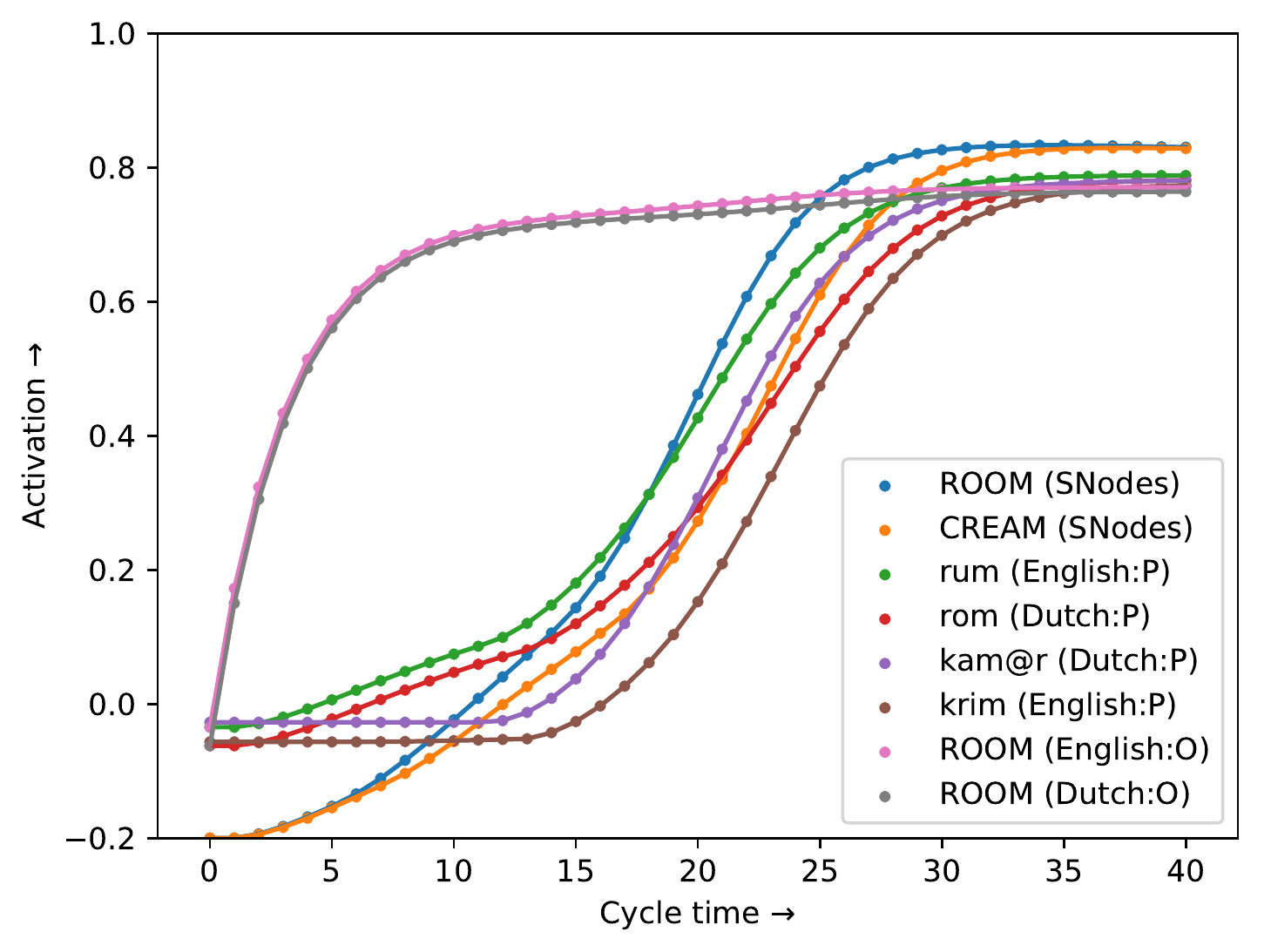}
        \caption{Matrix chart for input \texttt{ROOM} (cream$_{NL}$ or room$_{EN}$).}
        \label{fig:wt_matrix_room}
    \end{subfigure}
    \caption{Activation charts showing node activity simulated over time.}
    \label{fig:wt_matrix_plots}
\end{figure}

Let us turn to an example to illustrate the problem.
Consider the two activation charts in \fref{fig:wt_matrix_plots}.
On the left, we have presented the Dutch word \texttt{AARDBEI} to the model.
We first see the corresponding orthographic node become active.
Accordingly, the phonological node \texttt{ardbK} and semantic node \texttt{STRAWBERRY} start to become active.
Once the semantics are active enough, we finally see the English phonology \texttt{str\$b@rI} become active,
which the model selects for output after 32 time cycles.

Compare this to the activation chart for the interlingual homograph \texttt{ROOM} on the right.
Unlike \texttt{AARDBEI}, this word exists in both Dutch and English.
However, the Dutch word is equivalent to the English word `cream', not `room'.
As an orthographic representation is activated for both languages,
ultimately four phonetic representations become active,
two meaning `cream' and two meaning `room'.
This leads to a tight response competition process,
accompanied by selection problems.
As can be clearly seen in \fref{fig:wt_matrix_room},
all four phonological representations end up passing the 0.72 mark.
However, per the threshold criterion, only the first one passing this mark is selected.
In this case, \texttt{rum} is selected, while \texttt{krim} is expected.
How should we go about solving this problem?

\section{Proposed solution}

How interlingual homographs affect participant performance in word translation
has been the subject of considerable research.
Distinguishing different types of interlingual homographs,
\citeA{goertz2018} investigated these phenomena experimentally in a study of Dutch--English bilinguals.
Referring to initial Multilink simulations on the resulting data,
she proposes two additional selection mechanisms to facilitate interlingual homographs:
one set at the input level, and another one set at the semantic level.

At the input level, \citeauthor{goertz2018} proposes to introduce a \emph{shortlist} for activated words (pp.~45--46).
The items on the shortlist will be evaluated based on their associated language, starting with the most activated word.
If the first element on the list matches the target language, it is selected as the input node.
If not, the list is evaluated further until such a match is found.
A similar shortlist is proposed for the output (phonetic) candidate nodes.
Here, phonetic nodes passing an activation threshold of 0.72 will be evaluated,
based on whether their associated language matches the target language.
Finally, the output candidate is subjected to a \emph{semantic check} to ensure the input and output candidate have the same meaning.
If this is not the case, the next candidate in the shortlist will be evaluated instead, until such a match is found.

\begin{figure}[!t]
    \centering
    \includegraphics[width=0.95\linewidth,clip,trim=0.6cm 3.2cm 0.6cm 3.5cm]{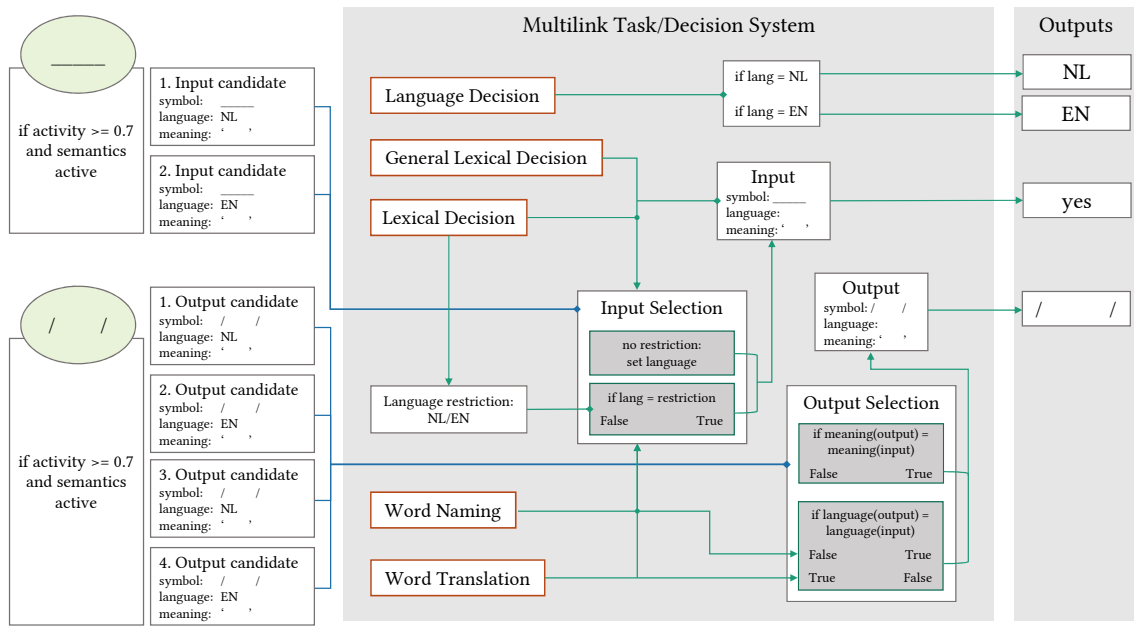}
    \caption{The proposed word translation task extensions \protect\cite{goertz2018} embedded in the task/decision system.
        Tasks are indicated in red, with blue lines for input, and green lines for process flow.}
    \label{fig:wt_extensions}
\end{figure}

The architecture of the revised cognitive control system is illustrated in \fref{fig:wt_extensions}.

These proposed changes to the model will solve the selection problem explained in the previous section.
However, in delaying output until the perfect candidate comes around, we make it harder to simulate \emph{human errors}.
These errors may depend on certain variables controlled for in experimental settings,
such as decision time allotted, task familiarity, etc.
Participant fatigue may increase the error rate as well, depending on task demands.

Simulating such errors is made more difficult by the current Multilink architecture being completely deterministic and focused on correctness.
Hence, for the moment, any simulations using the proposed mechanisms will only produce \emph{correct} responses,
if the nodes for the representations in question are available.
While this limits the simulations somewhat, we deem it necessary to first extend the model as proposed
before introducing any stochastic components to the model.

\section{Implementation}

The proposed solution to the interlingual homograph selection problem
has been implemented in Multilink in a new task in Multilink's task/decision system.
This new task makes use of shortlists to evaluate lexical candidates for input and output.
These shortlists are implemented as pool-level abstractions of the lexical network, and facilitate the evaluation of nodes.
These lists are analogous to \emph{waiting rooms}, in that new candidates come in over time, are evaluated,
and are removed if not applicable.
For reference, the main classes of the Java implementation have been included in appendices
\ref{app:trans_shortlist} and \ref{app:word_translation}.

Simulation work strongly suggests that lateral inhibition is insufficient to solve the homograph selection problem.
Therefore, as proposed, this new task specification holds that lexical candidates are explicitly checked with respect to
both language and semantics.
Unlike other task implementations, words may hence be rejected even if they pass the activation threshold,
based on these criteria of language membership and semantic equivalence.

\section{Initial findings}

\begin{figure}[!t]
    \centering
    \begin{subfigure}[b]{0.47\textwidth}
        \includegraphics[width=\textwidth]{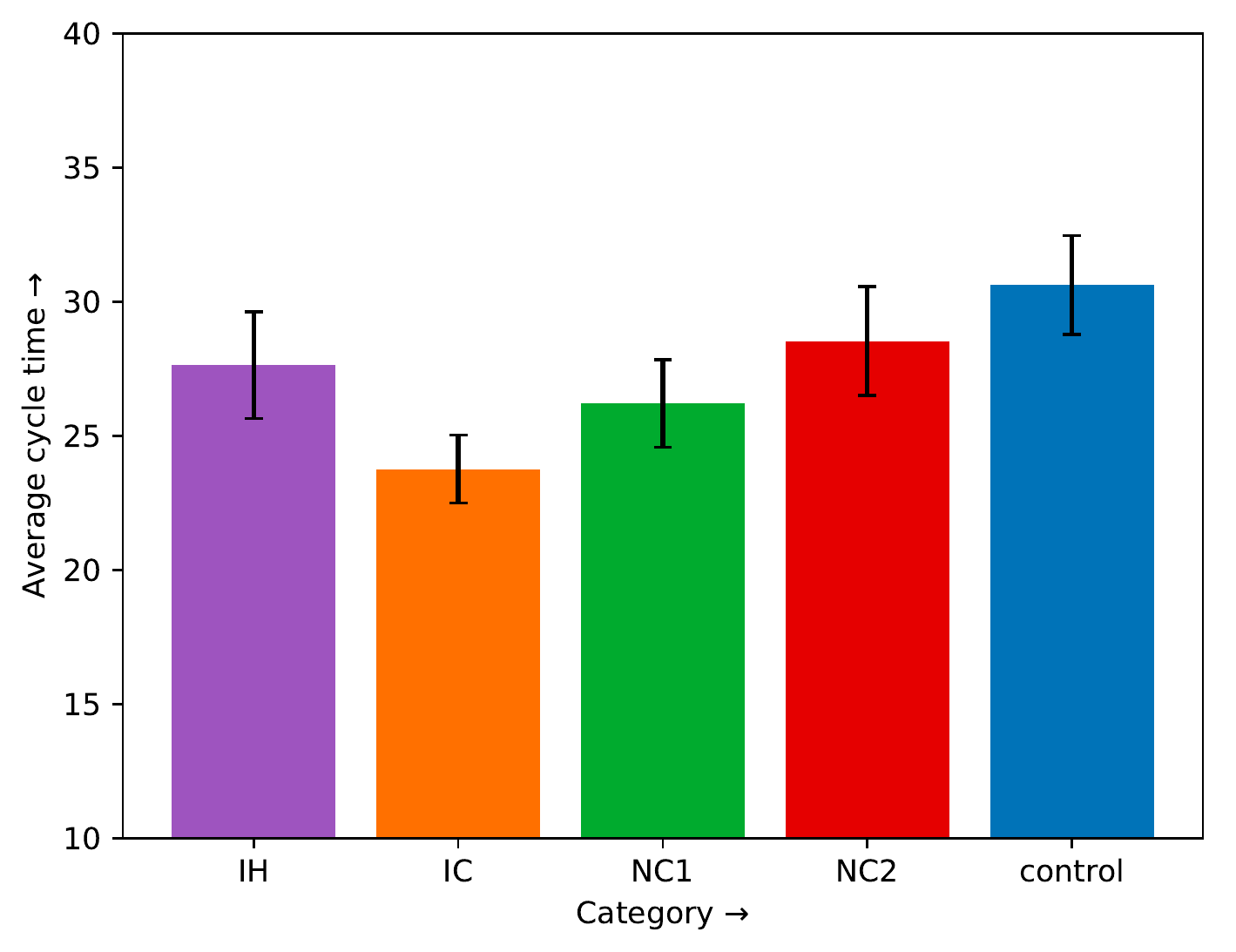}
        \caption{Recognition task simulation}
        \label{fig:res_bars_wt_R_with_LI}
    \end{subfigure}
    ~
    \begin{subfigure}[b]{0.47\textwidth}
        \includegraphics[width=\textwidth]{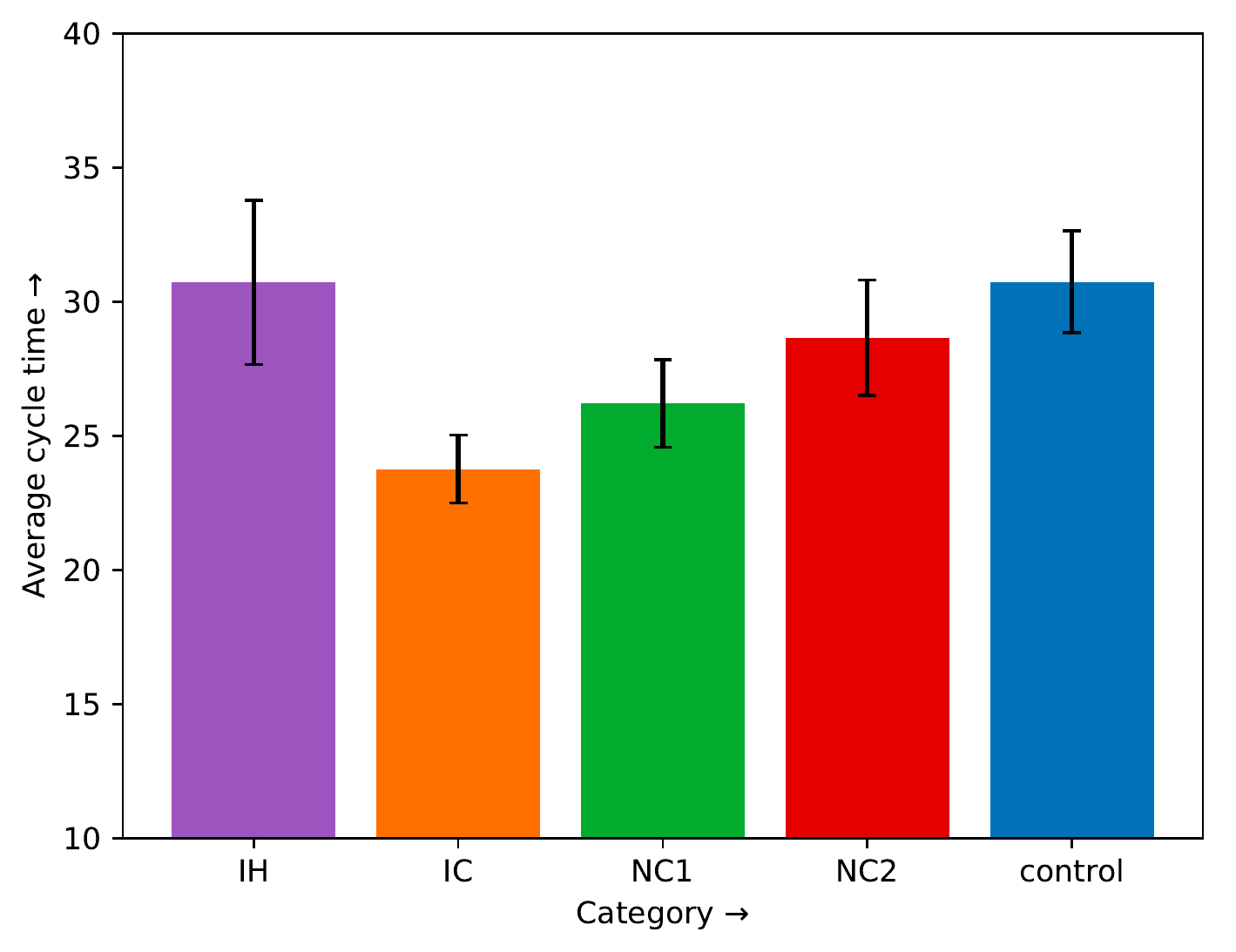}
        \caption{Word Translation task simulation}
        \label{fig:res_bars_wt_WT_with_LI}
    \end{subfigure}
    \caption{Average cycle times by category for a full-lexicon translation simulation of 1,466 English stimuli and Dutch targets.
        IH is the Interlingual Homographs category, whose cycle times now more closely resemble empirical data.
        IC indicates identical cognates, while NC1 and NC2 indicate non-identical cognates with a Levenshtein distance of 1 and 2, respectively.}
    \label{fig:wt_plots}
\end{figure}

To test the new word translation task implementation,
the Multilink lexicon was first extended with word pairs from the study by \citeA{goertz2018}.
Next, we performed a full-lexicon simulation of a translation production task for
the 1466 English stimuli and Dutch targets from the resulting lexicon.
As a baseline, we also performed a simulation with identical settings using the generic recognition task implementation.
The lateral inhibition parameters were $OO\gamma = PP\gamma = -0.001$ for both simulations.
For completeness, an overview of all parameter settings is included in \fref{app:parameters}.

The cycle times resulting from both simulations were averaged by word category.
The resulting averages are included as bar plots in \fref{fig:wt_plots}.
We note that, in the generic recognition task, interlingual homographs are incorrectly translated.
For example, the stimulus \texttt{ROOM} yields the phonological code \texttt{/rum/}$_{EN}$ (room) instead of
the translation \texttt{/krim/}$_{EN}$ (cream).
This results in lower cycle times for this category, which can be clearly seen in figure~\ref{fig:res_bars_wt_R_with_LI}.

For further validation,
we turn to another study on interlingual homograph recognition in Dutch--English participants \cite{vanlangendonckIP}.
When comparing the simulations to the empirical data at hand, we see an encouraging similarity in their distributions.
Compare figure~\ref{fig:res_bars_wt_WT_with_LI} to \fref{fig:res_vanlangendonckIP_pure}.
In both the simulation and the empirical study, interlingual homographs are found to be slowed down compared to
identical cognates, with control words being slightly faster.
Furthermore, non-identical cognates are processed slightly slower than the identical cognates,
although the empirical data show relatively less difference between these categories than the simulated data.

Importantly, we find that the word translation task we introduced, as well as the shortlists accommodating it,
results in correct translations for stimuli that did not do so in the generic recognition task.
The adaptation did lead to slower cycle times for the affected stimuli.
As hypothesised, this slowing of processing primarily affected interlingual homographs.

\section{Limitations}

\begin{figure}[!b]
    \centering
    \begin{subfigure}[b]{0.47\textwidth}
        \includegraphics[width=\textwidth]{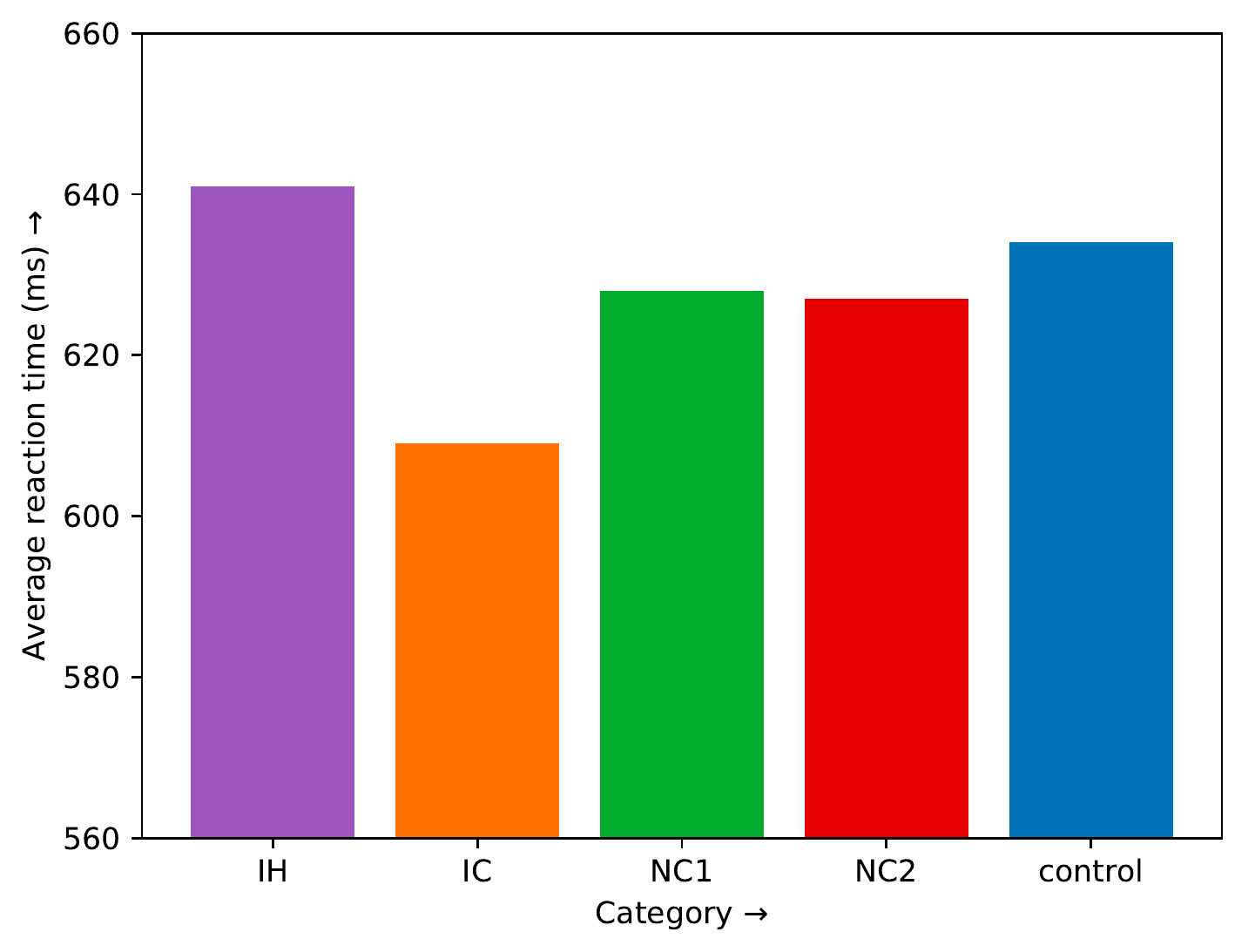}
        \caption{`Pure' condition, showing patterns similar to those in \fref{fig:res_bars_wt_WT_with_LI}}
        \label{fig:res_vanlangendonckIP_pure}
    \end{subfigure}
    ~
    \begin{subfigure}[b]{0.47\textwidth}
        \includegraphics[width=\textwidth]{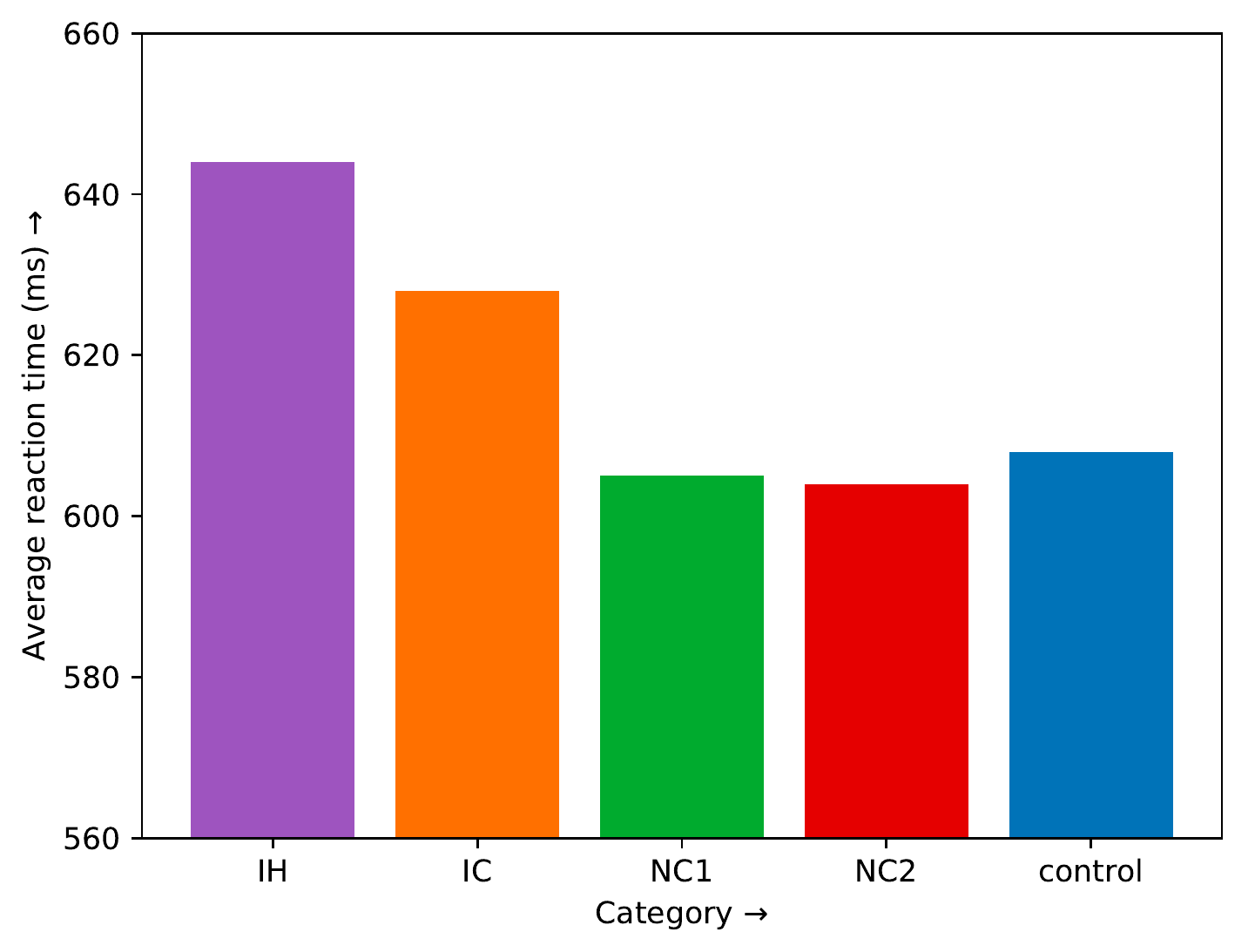}
        \caption{`Mixed' condition, showing reverse inhibitory patterns not yet simulated by Multilink.}
        \label{fig:res_vanlangendonckIP_mixed}
        \end{subfigure}
    \caption{Results from experimental Lexical Decision tasks \protect\shortcite{vanlangendonckIP}}
    \label{fig:bars_vanlangendonckIP}
\end{figure}

The study by \shortciteauthor{vanlangendonckIP} consisted of two experiments.
The first experiment tested words in a `pure' condition, consisting of only English words are pseudo-words.
In contrast, the second experiment added Dutch words to test a `mixed' condition.
Crucially, the authors found two very different reaction time patterns for these two conditions,
as depicted in \fref{fig:bars_vanlangendonckIP}.
As discussed, Multilink is able to account for the pattern found in the \emph{first} experiment.
Happily, a simulation of this experiment's stimuli resulted in an overall correlation $r = 0.69$
with the experiment's reaction times, averaged by item.
Unfortunately, we were not yet able to produce a similar success for the \emph{second} experiment.
An analogous simulation of experiment 2 resulted in a correlation $r = 0.44$ with empirical data.

We speculate that the difference in pattern arose because the second experiment's conditions
required participants to more explicitly verify the language membership of identical words.
This would result in a processing delay for both the IH and IC word categories,
while non-identical and control categories would profit from not having to check this membership.
The authors refer to this as a `mirrored inhibition effect'.

To simulate the combination of word retrieval and task/decision effects, the Multilink model
will have to be adapted and expanded.
We hypothesise that the solution will consist of a combination of four factors:
language node activation, a task specification that simulates actions based on this activation,
the $SP/PS_\alpha$ connection weights, and the $OO_\gamma$ inhibition setting.

In Multilink's lexical network, all orthographic and phonological nodes are connected to a language node.
Currently, these nodes are merely \emph{tags} that are not activated by their connections.
Because the connections do exist, however, a change in parameter settings (e.g. $LO_\alpha$) could activate them.
In a similar vein, language nodes could play a more active role in word competition via the $LO_\gamma$ and $LP_\gamma$
connections to \emph{inhibit} nodes based on membership of \emph{another} language.
Currently, this option, too, is disabled.

At present, word production is mediated by a boost between semantic and phonological nodes.
This boost was originally implemented to make sure that orthographic information to
the phonological nodes via semantics was strong enough to have noticeable effects.
The $SP/PS_\alpha$ boost allows phonological nodes, including translations, to become active within an acceptable time frame.
However, because phonological nodes are also connected to their orthographic counterparts through the $PO_\alpha$ connections,
which themselves are also connected to the language nodes, the boost indirectly affects the language nodes as well.
Therefore, it is advisable to re-evaluate the functionality and desirability of this boost before we simulate the findings from figure~\ref{fig:bars_vanlangendonckIP}.

\begin{figure}[!b]
    \centering
    \begin{subfigure}[b]{0.47\textwidth}
        \includegraphics[width=\textwidth]{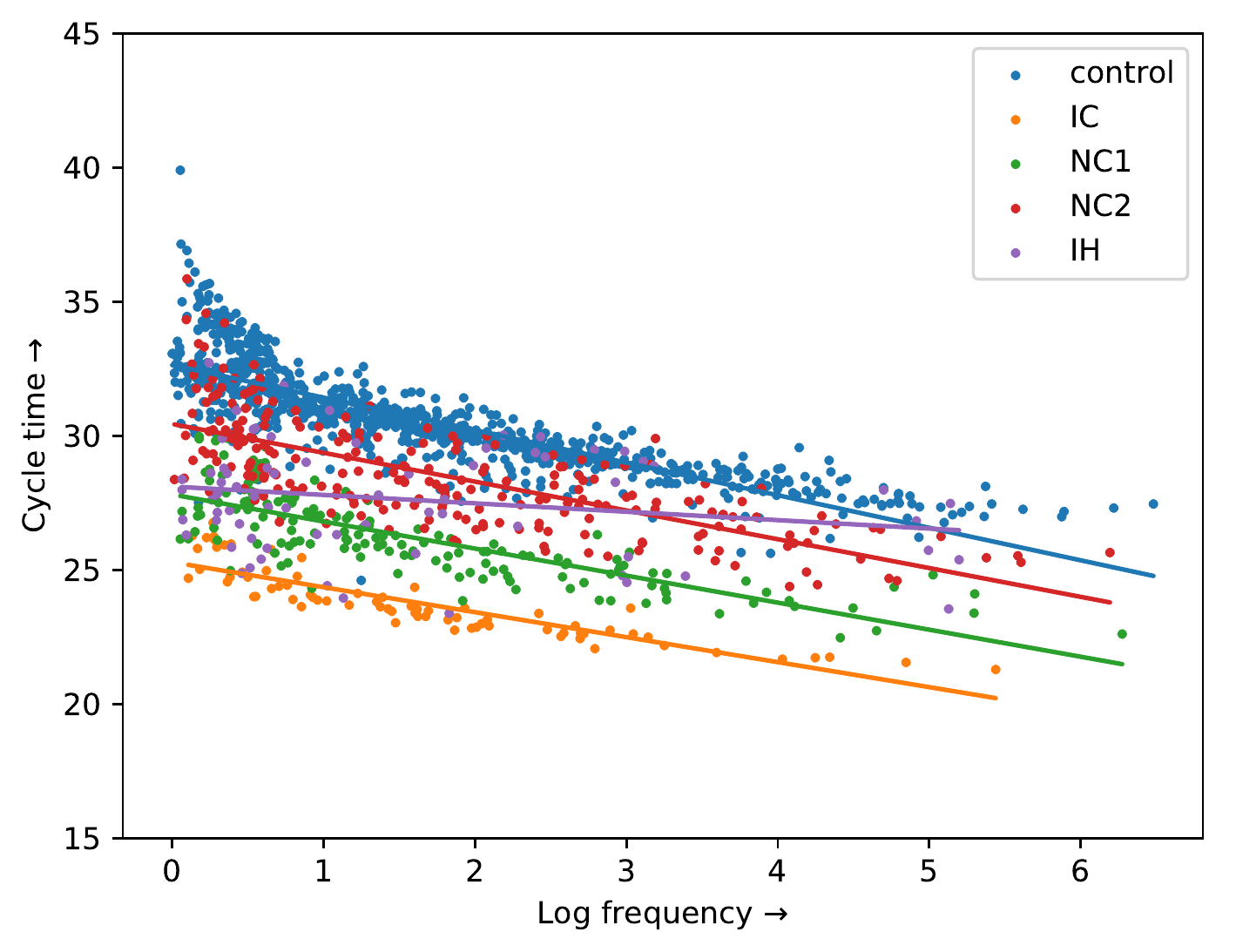}
        \caption{Recognition task simulation}
        \label{fig:scatter_R_with_LI}
    \end{subfigure}
    ~
    \begin{subfigure}[b]{0.47\textwidth}
        \includegraphics[width=\textwidth]{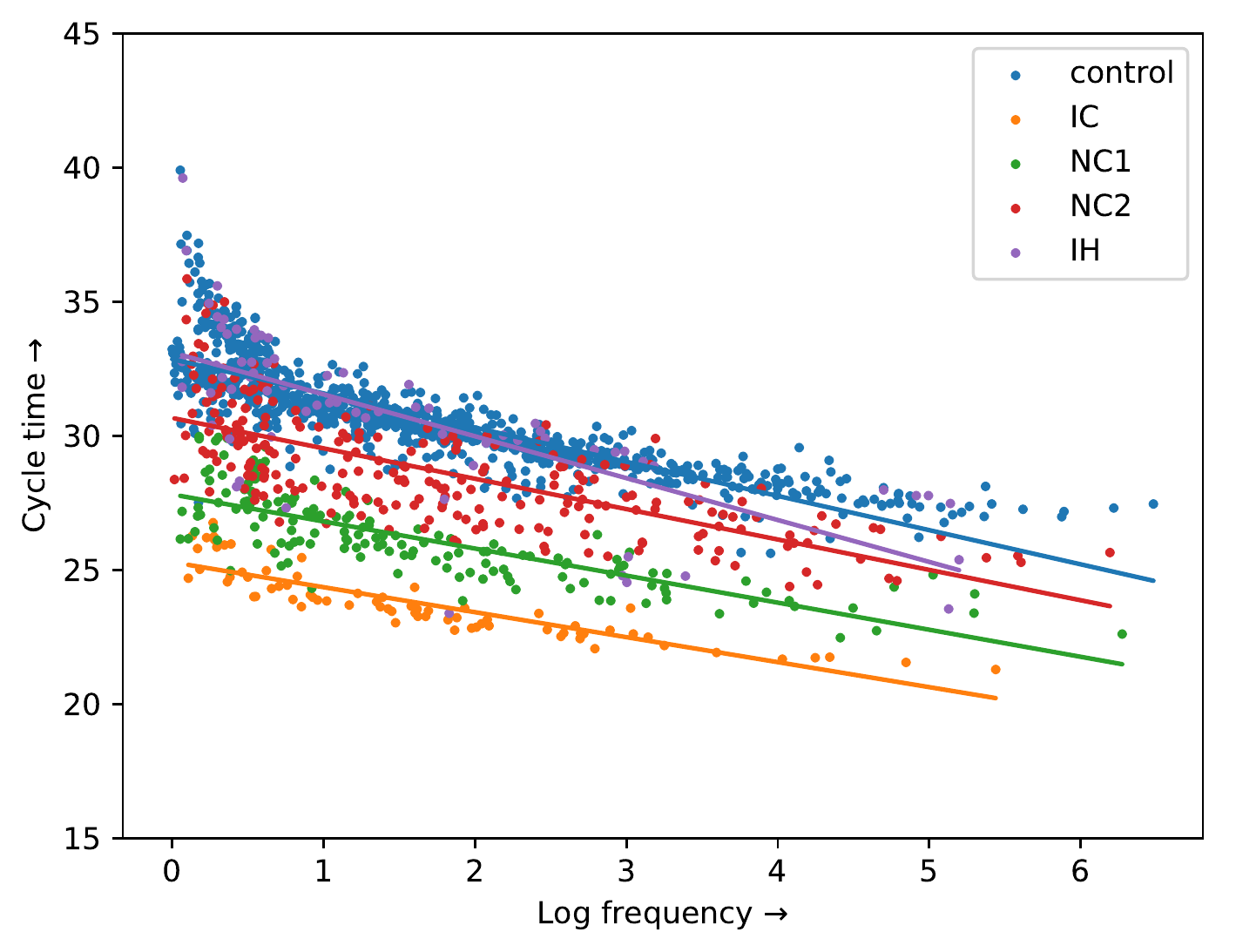}
        \caption{Word Translation task simulation}
        \label{fig:scatter_WT_with_LI}
        \end{subfigure}
    \caption{Results for full-lexicon word translation simulations of stimuli from the English Lexicon Project.
        Each data point corresponds with the retrieval of a phonological code by the model.}
    \label{fig:wt_scatter_plots}
\end{figure}

Finally, the $OO_\gamma$ inhibition setting controls the amount of inhibition between orthographic nodes,
regardless of language membership.
Preliminary findings suggest that a larger orthographic inhibition substantially increases the cycle times required to reach a decision
for words with dense neighbourhoods.
Hence, the setting of lateral inhibition may be a key factor in the solution as well.
Clearly, finding an optimal solution will require substantial exploration and effort.

\section{Conclusions}

We have seen how the processing of interlingual homographs poses special problems to Multilink for simulations on word translation tasks.
Our new implementation of a dedicated \emph{word translation} task addresses these problems and
offers more accurate simulations of this special category of words, resulting in more accurate reaction time patterns
than a generic word recognition task.

For validation purposes, we applied the new task implementation to the first experiment from \shortciteNP{vanlangendonckIP},
with a resulting $r = 0.69$. At the condition level, the observed simulation patterns provided a good match to the empirical data as well.
It was found that we currently could not yet simulate the second experiment from the same study with the same high degree of similarity.
Nevertheless, correlating the results for the simulation of the second experiment still resulted in a correlation $r = 0.44$.
We have laid out some ideas for implementation that may improve the accuracy for simulations like these,
but making these work requires substantial work yet.

In sum, while we can simulate the second \shortciteauthor{vanlangendonckIP} experiment to only limited extent,
we \emph{can} simulate the first experiment in a very satisfactory way.
Moreover, as we will see in the next chapter,
the current implementation is sufficient to simulate another study by \citeNP{goertz2018}, to high accuracy as well.

    \chapter{Translating Interlingual Homographs}

Word translation tasks pose an interesting challenge for participants in psycholinguistic studies.
Like lexical decision and word naming tasks, word forms will need to be retrieved from the mental lexicon.
The demands for a translation task a higher, however.
This becomes clear when we consider words that have some degree of form overlap between languages.

For \emph{cognates}, words with a high degree of both form and semantic overlap,
a consistent finding is that the translation process is sped up compared to control words.
This is known as the \emph{cognate facilitation effect} (e.g. \citeNP{christoffels2006, degroot1994forward}).
Conversely, if there is word form overlap, but no semantic overlap,
the translation process is \emph{slowed down} compared to control words.
This is known as the {interlingual homograph interference effect}
\shortcite{dijkstra1998interlingual, christoffels2013language}.
This effect has been found to depend on whether another language is involved in the task \shortciteA{dijkstra1998interlingual}.
In a task with purely monolingual demands, \shortciteauthor{dijkstra1998interlingual} find the effect is absent,
while it does occur with the same participants in a bilingual context.

These findings, in turn pose a challenge for the computational modelling of language processes.
Like a participant, a model has to make a decision with respect to what output to produce.
For lexical decision tasks, we find it suffices to respond to certain word form activation only.
However, for word translation, in particular interlingual homographs,
this may leads to a response based on the wrong word form representation.
This is problematic, as this then produces an incorrect response.
In turn, this response is faster than is typically found in empirical data.

In order to ensure a retrieved word form is the correct translation of another,
a solution could be to introduce an explicit check that compares semantics
for the respective input and output candidates.
We discussed a solution for these word selection problems in chapter~\ref{ch:word_translation},
and analysed its effect on a full-lexicon simulation.

We put the proposed mechanism to the test by simulating an extensive empirical study by \citeA{goertz2018}.
This study investigated these word selection problems in a word translation task with proficient Dutch--English bilinguals.
\citeauthor{goertz2018} kindly provided us with the raw trial data from her experiment
(N=7,696, excluding practice trials and fillers).
The experimental means per item and per condition were computed based on the steps taken in
the original analysis \cite[pp. 21--22]{goertz2018}.
We used Python 3.7.4 combined with the Pandas 0.25.1 package for this process.
After clean-up, the final dataframe contained 5,304 trials (68.91\%).

In the following sections, we will run Multilink simulations for the same stimulus set and then compare
the model's simulation results to the empirical data from \citeA{goertz2018}.
As we will see, Multilink is able to replicate the behavioural patterns on multiple test conditions,
including those that were beyond reach before (e.g, interlingual homographs).
Thus, Multilink can be argued to provide adequate explanations of the bilingual word selection process,
which can be attributed to the newly introduced word translation mechanism.

\section{Cognate effects in IH translation}

The first goal of \citeauthor{goertz2018}'s study was to reproduce the cognate facilitation
and IH interference effects by manipulating the cognate status of the word stimuli.
Proficient Dutch--English bilinguals were asked to pronounce the translation of a word presented on screen as quickly and correctly as they could.
Items were presented in two blocks, each requiring translation in one direction (i.e. from Dutch to English or vice versa).

The following four item--target conditions were distinguished in the first part of the analysis \cite[pp. 23--30]{goertz2018}.

\begin{description}
    \item[Interlingual Homograph (IH)] The input item and translation have no orthographic overlap, but a word exists
        in the target language \emph{with} full orthographic overlap \emph{and} a completely different meaning.
        Example: \texttt{RAGE}$_{NL}$--\texttt{TREND}$_{EN}$.
    \item[IH/Cognate] The input item and translation have a high degree of orthographic overlap, but a word exists
        in the target language with \emph{full} orthographic overlap \emph{and} a completely different meaning.
        Example: \texttt{BOOT}$_{NL}$--\texttt{BOAT}$_{EN}$.
    \item[Control] The input item and its translation have no orthographic overlap, nor does the input item occur in the target language \emph{as-is}.
        Example: \texttt{FIETS}$_{NL}$--\texttt{BIKE}$_{EN}$.
    \item[Control/Cognate] Input and translation have a high degree of orthographic overlap. The input item does not occur in the target language \emph{as-is}.
        Example: \texttt{SIGAAR}$_{NL}$--\texttt{CIGAR}$_{EN}$.
\end{description}

\begin{figure}[!b]
    \centering
    \begin{subfigure}[b]{0.49\textwidth}
        \includegraphics[height=5.5cm]{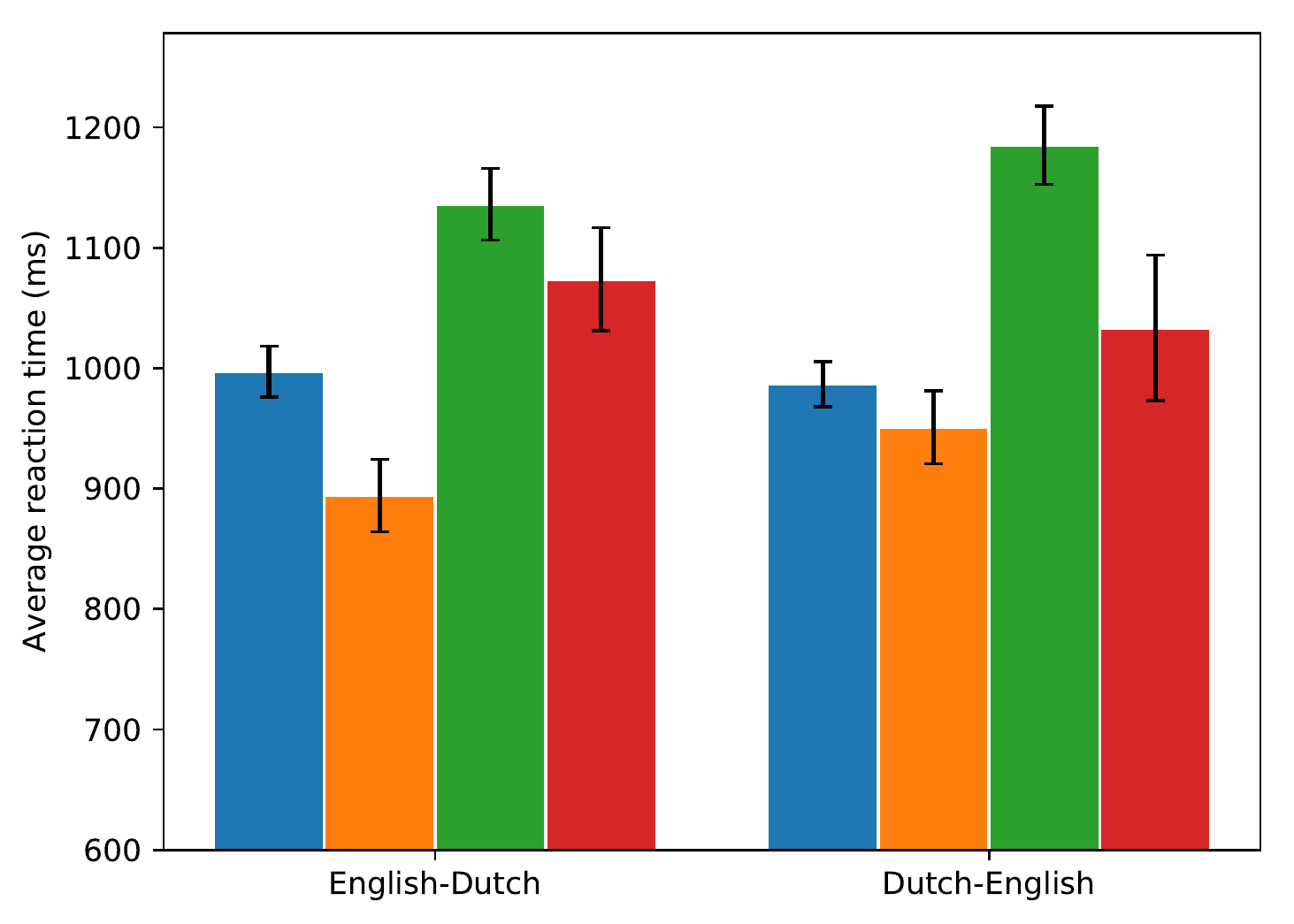}
        \caption{Average reaction times by condition.}
        \label{fig:goertz_ff_RTs}
    \end{subfigure}
    ~
    \begin{subfigure}[b]{0.49\textwidth}
        \includegraphics[height=5.5cm]{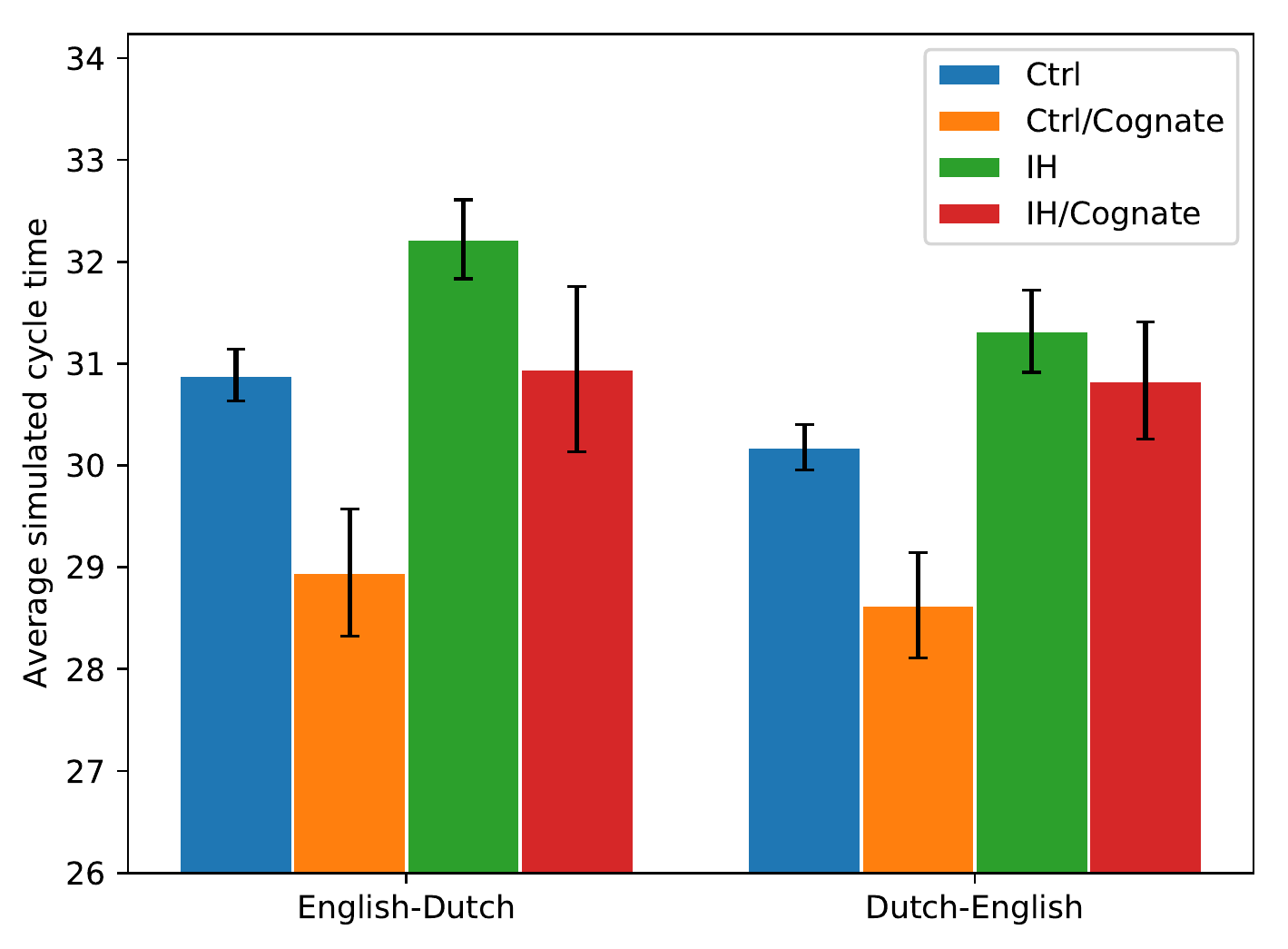}
        \caption{Simulated reaction times by condition.}
        \label{fig:goertz_ff_cycles}
    \end{subfigure}
    \caption{Reaction times from \protect\citeA{goertz2018} word naming task involving interlingual homographs,
        split by condition (left). Multilink produces cycle times of the same pattern (right) with $OO_\gamma = -0.03$.}
    \label{fig:goertz_ff}
\end{figure}

Stimuli from these conditions were used as inputs to Multilink.
As \fref{fig:goertz_ff} illustrates,
Multilink was able to produce cycle times that exhibit the same pattern as the empirical data from \citeA{goertz2018}.
These included a cognate facilitation effect, as well as slower response times when interlingual homographs were involved.

Next, the role of lateral inhibition in terms of the fitness of model to data was extensively investigated
by varying Multilink's $OO_\gamma$ and $PP_\gamma$ hyperparameters.
Simulated cycle times were correlated with reaction times from \citeA{goertz2018}.
The results are summarised in \fref{tab:goertz_ff} above.
For both model variants listed, the $PP_\gamma$ parameter was set to 0.0.
This setting was found to yield a better fit, regardless of the $OO_\gamma$ setting.
However, this implies that phonological inhibition was disabled in the resulting models.
This setting seems counter-intuitive for a model of the translation production task,
in which phonological representations must be retrieved to utter the translated words.
Why does this then result in an apparently optimal model?

Recall that we discussed the number of active nodes over time depending on the strength of lateral inhibition
(cf. \fref{fig:fit_li_num_nodes}).
For the present study, we found substantially improved correlations for a \emph{higher} orthographic inhibition value
compared to the base inhibition value established in chapter~\ref{ch:li_fitting}.
However, the phonological inhibition was found to have \emph{too much} of an inhibiting effect on the network.
We explain this effect by the small number of phonological nodes that are active
when such a higher value for the orthographic inhibition parameter is used.
In such cases, the orthographic nodes \emph{compete} for activation,
in turn leading to \emph{fewer} active phonological nodes.
This account explains why enabling (strong) competition among phonological nodes appears to have a detrimental effect on node selection.
Therefore, we conclude a good fit is dependent on sufficient, but not too much, lateral inhibition.

\begin{table}[!t]
    \centering
    \begin{tabular}{llrrrrr}
        \toprule
                   &                       &            & \multicolumn{2}{c}{$OO_\gamma = -0.03$} & \multicolumn{2}{c}{$OO_\gamma = -0.0001$} \\
        \cmidrule(r){4-5}
        \cmidrule(r){6-7}
        Direction  &  Condition            &   Avg. RT  & Avg. cycle  &  Correlation  & Avg. cycle  &  Correlation \\
        \midrule
        D-En       &  Control              &   986.772  &     30.179  &          0.298   &     29.394  & \textbf{0.314} \\
                   &  Control/Cognate      &   950.817  &     28.627  &  \textbf{0.365}  &     26.537  &         0.292  \\
                   &  IH                   &  1184.995  &     31.318  &  \textbf{0.581}  &     30.065  &         0.544  \\
                   &  IH/Cognate           &  1033.409  &     30.833  &  \textbf{0.538}  &     27.347  &         0.366  \\
        \midrule
        En-D       &  Control              &   997.155  &     30.886  &          0.314   &     30.132  & \textbf{0.423} \\
                   &  Control/Cognate      &   894.057  &     28.949  &          0.512   &     26.637  & \textbf{0.520} \\
                   &  IH                   &  1136.150  &     32.220  &  \textbf{0.613}  &     30.731  &         0.548  \\
                   &  IH/Cognate           &  1073.800  &     30.947  &  \textbf{0.569}  &     27.377  &         0.331  \\
        \midrule
        D-En       &  overall      \hfill (N=127) &  1078.730  &     30.398  &  \textbf{0.534}  &     28.972  &         0.451  \\
        En-D       &  overall      \hfill (N=127) &  1059.575  &     30.983  &  \textbf{0.570}  &     29.368  &         0.478  \\
        Overall    &  by item      \hfill (N=254) &  1068.495  &     30.690  &  \textbf{0.538}  &     29.171  &         0.455  \\
        Overall    &  by condition \hfill (N=8)   &  1032.145  &     30.495  &  \textbf{0.853}  &     28.528  &         0.639  \\
        \bottomrule
    \end{tabular}
    \caption{Results from correlating empirical reaction times from \protect\citeA{goertz2018} with predictions from Multilink simulations.}
    \label{tab:goertz_ff}
\end{table}

\section{Hidden cognate effects in IH translation}

\citeauthor{goertz2018} found that interlingual homographs that were also cognates with respect to the input language
(e.g. \texttt{BOOT}$_{NL}$--\texttt{BOAT}$_{EN}$) were processed significantly faster than interlingual homographs without a hidden cognate.
Thus, there are effects of the non-targeted reading of the stimulus on processing.
This can be explained by assuming that the Dutch reading of the word is competing with the English reading, and vice versa.
In the process, cognates are facilitated due to semantic overlap, while IHs are slowed due to response competition.
This hinders participants in quickly naming the correct translation.

\clearpage
Three conditions were considered in the second part of the analysis \cite[pp. 31--39]{goertz2018}:

\begin{description}
    \item[Interlingual Homograph (IH)] The input item and translation have no orthographic overlap, but a word exists
        in the target language \emph{with} full orthographic overlap \emph{and} a completely different meaning.
        Example: \texttt{RAGE}$_{NL}$--\texttt{TREND}$_{EN}$.
    \item[IH/Cognate] The input and translation have a high degree of orthographic overlap, however a word exists
        in the target language with \emph{full} orthographic overlap \emph{and} a completely different meaning.
        Note that is intended to hinder both \emph{item} selection and \emph{target} selection.
        Example: \texttt{BOOT}$_{NL}$--\texttt{BOAT}$_{EN}$.
    \item[IH/Hidden] The input and translation have a high degree of orthographic overlap. However, the target language
        contains a word of completely different meaning \emph{with} full orthographic overlap with another word in the item's language.
        Note that is intended to hinder \emph{input item} selection instead of \emph{target} selection.
        Example: \texttt{ANGEL}$_{NL}$--\texttt{STING}$_{EN}$.
\end{description}

\begin{figure}[!b]
    \centering
    \begin{subfigure}[b]{0.49\textwidth}
        \includegraphics[height=5.5cm]{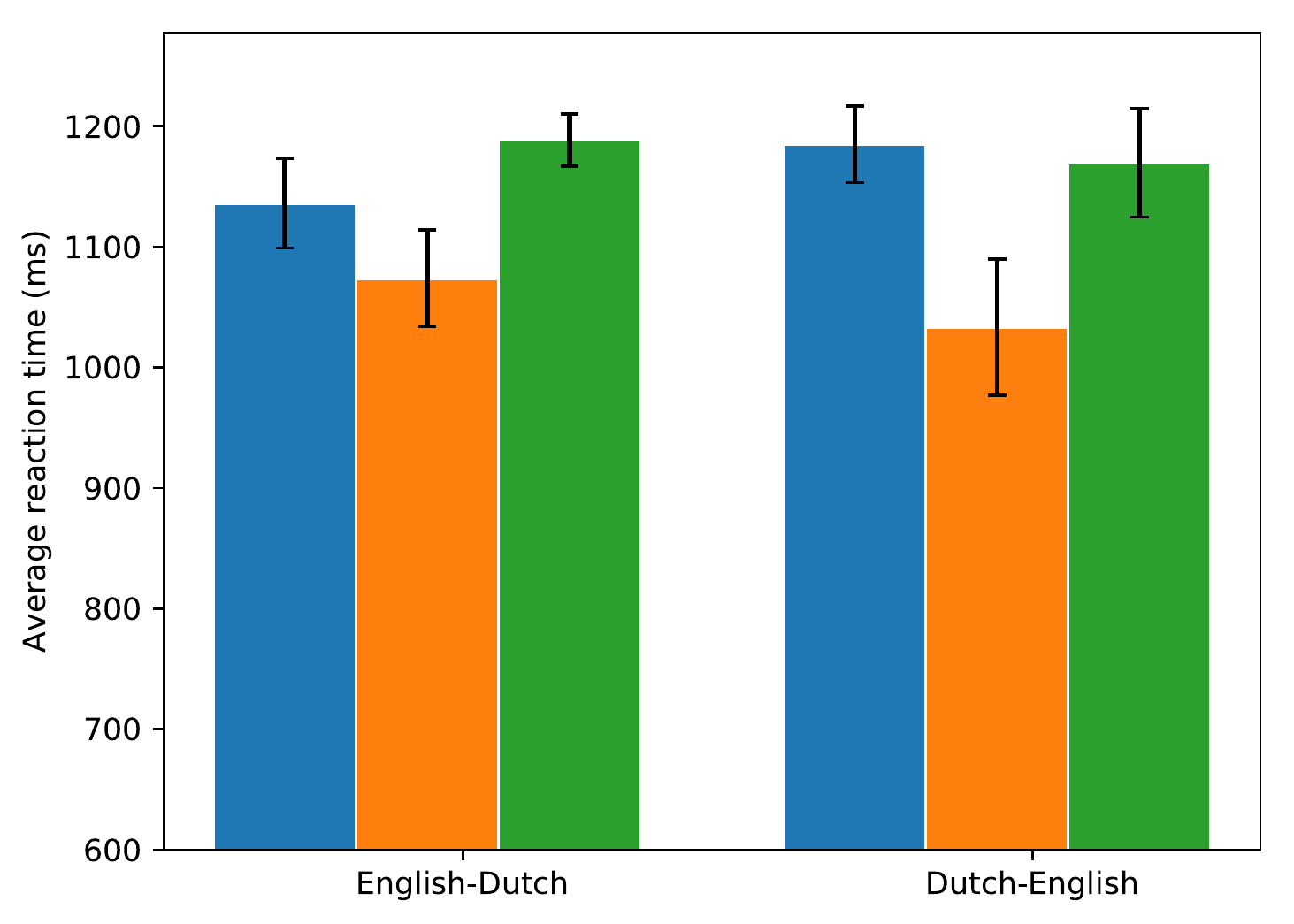}
        \caption{Average reaction times by condition.}
        \label{fig:goertz_hidden_RTs}
    \end{subfigure}
    ~
    \begin{subfigure}[b]{0.49\textwidth}
        \includegraphics[height=5.5cm]{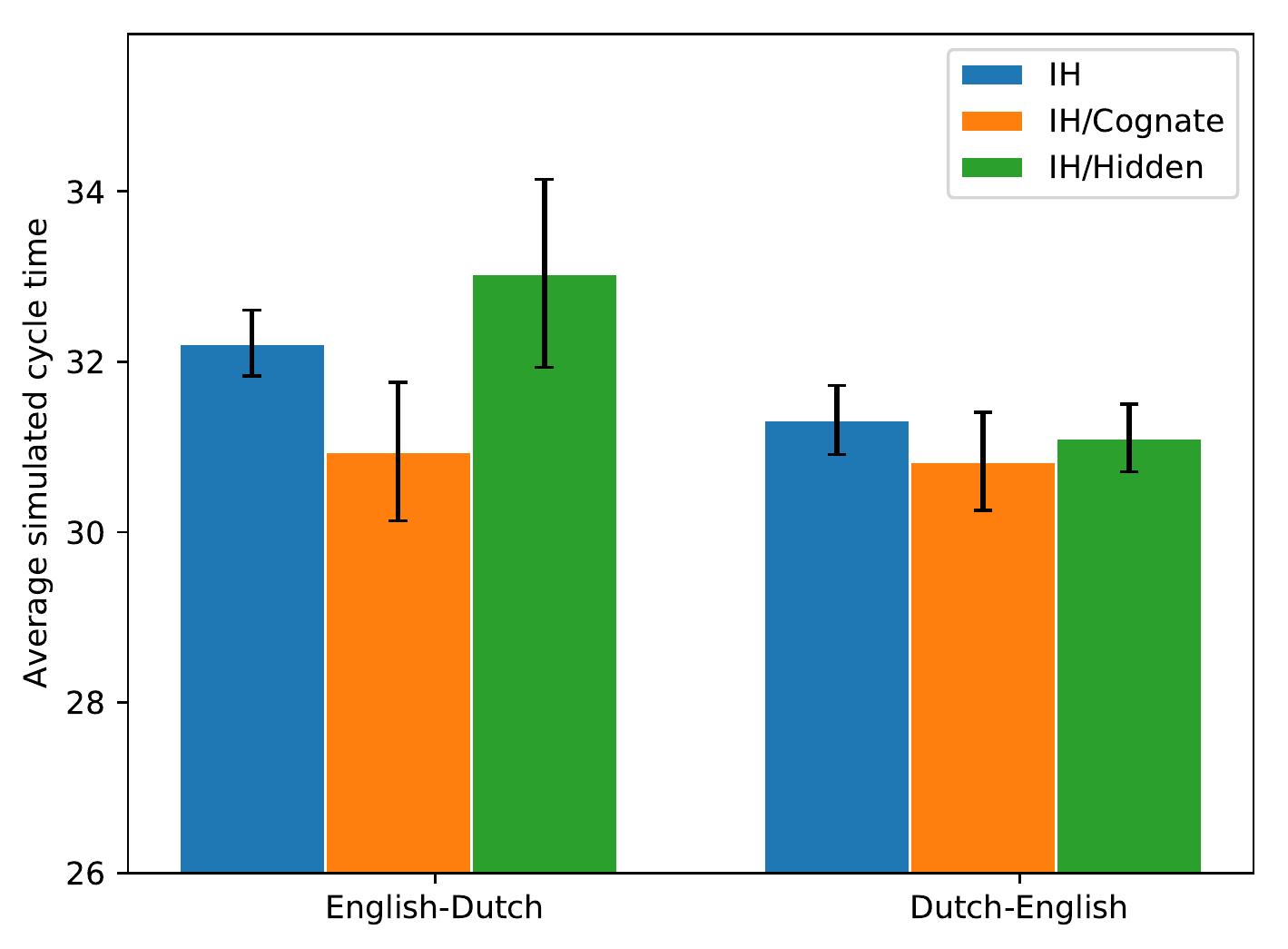}
        \caption{Simulated reaction times by condition.}
        \label{fig:goertz_hidden_cycles}
    \end{subfigure}
    \caption{Reaction times from \protect\citeA{goertz2018} showing hidden cognate effects in word naming task
        involving interlingual homographs (left). Multilink simulations come close with $OO_\gamma = -0.03$ (right).}
    \label{fig:goertz_hidden}
\end{figure}

The stimuli for each of these conditions were, once more, used as inputs to Multilink.
In line with the findings discussed in the previous section,
the inhibition settings $OO_\gamma = -0.03$ and $PP_\gamma = 0.0$ were used.

Analysis of the resulting cycle times reveals a pattern similar to the empirical data from \citeA{goertz2018}.
\Fref{fig:goertz_hidden} illustrates the reaction time means by translation direction and condition.
We find that the resulting patterns are very pronounced in the English--Dutch translation direction,
both for the empirical data and the simulations.
However, for the Dutch--English translation direction,
the patterns are much less pronounced in the simulations compared to their real-world counterparts.
Why is this?

To investigate this question, we returned to the original setting for lateral inhibition and
presented the same set of stimuli to the model. The resulting cycle times were averaged by condition.
We found that the IH/hidden condition yielded higher correlations for $OO_\gamma = -0.0001$,
but only in the Dutch--English translation direction.
Notably, we found that certain word stimuli were recognised in the $OO_\gamma = -0.0001$ condition,
but not in the $OO_\gamma = -0.03$ condition.
Remarkably, the same is true \emph{vice versa}.

\Fref{tab:goertz_hidden} shows the means for cycle and reaction times, as well as the correlations by translation direction and condition
for both of the $OO_\gamma$ settings.

\begin{table}[!t]
    \centering
    \setlength\tabcolsep{4.5pt} 
    \begin{tabular}{llrrrrrrrr}
        \toprule
                   &               & \multicolumn{4}{c}{$OO_\gamma = -0.03$} & \multicolumn{4}{c}{$OO_\gamma = -0.0001$} \\
        \cmidrule(r){3-6}
        \cmidrule(r){7-10}
        Direction  & Condition     &    N  &  Avg. RT  &          Corr. & Avg. cycle &   N  &  Avg. RT  &          Corr. & Avg. cycle \\
        \midrule
        D-En       & IH            &   37  & 1184.995  & \textbf{0.581} &     31.318 &  37  & 1184.995  &         0.544  &     30.065 \\
                   & IH/cognate    &   16  & 1033.409  & \textbf{0.538} &     30.833 &  16  & 1033.409  &         0.366  &     27.347 \\
                   & IH/hidden     &   16  & 1169.915  &         0.670  &     31.104 &  17  & 1169.915  & \textbf{0.768} &     30.494 \\
        \midrule
        En-D       & IH            &   36  & 1136.150  & \textbf{0.613} &     32.220 &  35  & 1136.150  &         0.548  &     30.731 \\
                   & IH/cognate    &   18  & 1073.800  & \textbf{0.569} &     30.947 &  19  & 1073.800  &         0.331  &     27.377 \\
                   & IH/hidden     &   13  & 1188.595  & \textbf{0.695} &     33.041 &  14  & 1188.595  &         0.622  &     32.084 \\
        \midrule
        D-En       & overall       &   69  & 1197.636  &         0.573  &     31.156 &  70  & 1204.965  & \textbf{0.609} &     29.548 \\
        En-D       & overall       &   67  & 1166.879  & \textbf{0.634} &     32.037 &  68  & 1165.010  &         0.530  &     30.073 \\
        Overall    & by item       &  136  & 1182.484  & \textbf{0.570} &     31.590 & 138  & 1185.277  &         0.545  &     29.807 \\
        Overall    & by condition  &    6  & 1131.144  &         0.586  &     31.577 &   6  & 1131.144  & \textbf{0.902} &     29.683 \\
        \bottomrule
    \end{tabular}
    \caption{Results from correlating empirical reaction times from \protect\citeA{goertz2018} with predictions from Multilink simulations.
        Note that Multilink is able to account for the inhibited cognate effect as well.}
    \label{tab:goertz_hidden}
    \setlength\tabcolsep{6pt} 
\end{table}

\section{Conclusions}

A simulation of the study by \citeA{goertz2018} showed that the Multilink model is able to simulate not only
the observed cognate facilitation effect (e.g. \texttt{SIGAAR}$_{NL}$--\texttt{CIGAR}$_{EN}$),
but the interlingual homograph interference effect as well (e.g. \texttt{RAGE}$_{NL}$--\texttt{TREND}$_{EN}$).
In short, Multilink can account not only for the cognate facilitation and IH interference effects,
but even for their combination.
This is quite remarkable, as these are emergent properties that result of a system consisting of relatively simple rules.

The key ingredients to these successes are the implementation of lateral inhibition and the new word translation task demands.
The former aspect has introduced active word competition to the model, resulting in the desired inhibition effects
observed in this chapter.
The latter aspect has introduced a set of rules allowing the model to select the correct output representation
when numerous alternatives are available in a manner that cannot be resolved through competition alone.

Importantly, we observed that different task situations called for different degrees of inhibition in the model.
While this conclusion warrants further research, it shows that lateral inhibition is a powerful mechanism
to influence model--data fitness.
Indeed, what we have so far referred to as `lateral inhibition' could perhaps be separated into
a more general network component (pertaining to lexical competition) as well as
several task-specific components (pertaining to decision and response competition).

    \chapter{General Discussion}

Over the course of this thesis, we have discussed several extensions to the Multilink model,
and the impact these have on model fitness.

Perhaps most notable is the introduction of \emph{lateral inhibition} to the model's lexical network.
This word competition mechanism allows word form representations compete for activation.
We have fit the hyperparameters for this new competition mechanism to reaction time data
from three extensive lexical decision studies:
the English Lexicon Project \shortcite{balota2007english}, the British Lexicon Project \shortcite{keuleers2012british},
and the Dutch Lexicon Project \shortcite{keuleers2010dutch}.
As a result, we found an optimal, generalisable parameter set for the lateral inhibition parameters
$OO_\gamma = PP_\gamma = -0.0001$.
In studying the number of active nodes in the model,
we find this parameter set reduces background activity by as much as 30\%.
As a result of this noise reduction, simulations were found to be faster to compute.

In addition to reducing noise in the model,
we have found this mechanism to be essential in reproducing \emph{delays} in the translation process
dealing with interlingual homographs (e.g. \texttt{ROOM}$_{NL}$--\texttt{CREAM}$_{EN}$).
Like cognates, this category of words has a high degree of form overlap between languages.
However, unlike cognates (e.g. \texttt{TUNNEL}$_{NL}$--\texttt{TUNNEL}$_{EN}$),
the meaning of these words is completely different.
As a result, this causes delays in the decision process in participants.
This effect is known as the \emph{interlingual homograph interference effect}.

The added competition mechanism alone is not enough to accurately predict the outcome of
word translation production tasks involving interlingual homographs.
We found the model would select a wrong output candidate most of the time.
For instance, given the stimulus \texttt{ROOM}, both the Dutch word \texttt{ROOM}$_{NL}$ and
the English word \texttt{ROOM}$_{EN}$ become active in the lexical network.
As a result, so do their respective semantics and phonological representations.
In such instances, four phonological representations become highly active.
The model would immediately select the first phonological representation to pass the activation threshold.
More often than not, it would select the wrong representation.
In this instance, it would choose \texttt{/rum/} (room) rather than \texttt{/krim/} (cream).

To account for these tasks, we have argued a check over semantics is required.
This was implemented in Multilink as a new task description in the model's \emph{task/decision system}.
Like a simulated Lexical Decision task (LD), this new simulated Word Translation task (WT) checks input activation
against a threshold in order to ascertain the input word.
However, unlike the LD task, output candidates meeting threshold are now explicitly checked to match the input's semantics.
If semantics do not match, the candidate is rejected from the output shortlist.
This, then, eventually results in selecting the right candidate once it has becomes active enough.
In the previous example, where the stimulus \texttt{ROOM} is presented to the model,
the candidate \texttt{/rum/} (room) is still considered first.
However, it is rejected on the basis of not matching the input semantics.
The candidate \texttt{/krim/} (cream) is considered next, and accepted.
As a result, the right response is returned.

\section{Future work}

In this thesis, we have extended Multilink in several ways.
The lack of lateral inhibition in the model was a recurring point of criticism to the model,
and hence made it to the research agenda \cite[708--709]{dijkstra2019modelling}.
Happily, this has now been implemented.
In a similar vein, some of the issues surrounding cognitive control (pp. 706--707)
have been resolved in our implementation of the new Word Translation task.
However, new issues have come to light as a result of our work on these extensions,
and old issues from the research agenda have been highlighted once more.
We will discuss these here.

\subsection{Lateral inhibition}

With lateral inhibition added to the model, we have added a mechanism with which
words compete in the lexical network, proportional to their activation.
Fitting this mechanism on extensive lexical decision studies revealed two optima:
one around the -0.0001 mark, and one around the -0.4 mark.
The latter is useful for fast simulations, but not generalisable to other tasks,
as all competitors, including translations, are inhibited.
We initially found the -0.0001 setting to generalise promisingly across conditions.

However, issues arose when simulating the study by \citeA{goertz2018}.
We found a lateral inhibition setting of -0.0001 to favour cognates too much
over conflicting interlingual homographs.
Instead, an optimal setting of -0.03 was found for simulations surrounding this study.
Interestingly, this setting is identical to the other connection weights in the model,
excluding the boosted connections between semantics and phonological nodes.

This does give cause to further research, however.
In the last keynote paper \cite{dijkstra2018multilink},
only one set of parameters was used to simulate all of the studies discussed.
Our findings here suggest there may be a task-specific component to lateral inhibition.
We suggest that what we have so far referred to as `lateral inhibition'
could perhaps be separated into a more general network component (pertaining to lexical competition)
as well as several task-specific components (pertaining to decision and response competition).

\subsection{Role of language nodes}

In this thesis, we have consistently compared the condition patterns emerging from
simulations to those found in experimental data.
Testing the new word translation task, we performed a full-lexical translation simulation.
The resulting condition patterns were compared to a study involving both cognates
and interlingual homographs \cite{vanlangendonckIP}.
We found that Multilink provides a good fit for the data from the first experiment from the
\shortciteauthor{vanlangendonckIP} study.
However, the second experiment, with heavier task demands, shows inhibitory patterns that we cannot yet reproduce.
We have suggested the answer may lie in the role of the language nodes.

Currently, Multilink uses language nodes as a membership `tag' only;
unlike in e.g. the BIA model, they exert no influence on the activation of nodes.
This has thus far proven to provide a good model-to-data fit.
However, the lack of inhibition from one language node unto nodes belonging to other languages
provides an interesting explanation for Multilink not being able to reproduce the second experiment from the
\shortciteauthor{vanlangendonckIP} study.
Therefore, we propose to explore whether activating the gamma connections involved,
in a way similar to the BIA model \cite[p. 475]{vanheuven1998},
could lead to the desired patterns.
Finding the right parameter settings may be a challenge, however.

\subsection{Semantic mediation}

Multilink's lexical network initialises its nodes using a \emph{resting-level} activation.
For orthography and phonology, this resting-level activation is based on word frequency.
However, the activation of semantics was not made dependent on the frequency with which a (lexical) concept is used.
Instead, the resting-level activation is set to the minimal activation, -0.2, for all semantic nodes.
As a result, the activation of semantics is guided only by the activity of
orthography and phonology, and not by any conceptual differences.
Whether or not some concepts are easier to access than others has not been subject to much research yet.
It could be that adding a frequency-dependence to concept activation would improve simulation results.
Indeed, making conceptual activation frequency-dependent is relatively easy to implement,
and could potentially solve another issue, namely the following.

Currently, activation of phonological nodes is heavily influenced by semantics.
Indeed, as we have seen, this is an essential part of the word translation process.
However, unlike other connections in the model, the connections between semantics and phonology
(SP/PS connections) are boosted.
While other alpha connections typically use a weight of 0.03,
these SP/PS connections use a weight of 0.3.
In essence, this means they propagate their activation 10 times as much.
Extrapolating from the simulation cycles required to perform lexical decision,
this boost was originally introduced as a means to meet the time frames found in experimental translation conditions.
Hypothetically, however, moving resting-levels for semantics to activating based on conceptual frequency
could reduce the need for such a boost, or perhaps make it entirely unnecessary.

\subsection{Non-alphabetic simulations}

However, in experimental settings, cognate facilitation effects have been found even across scripts (e.g. \citeNP{miwa2014}).
In this thesis, we have only considered simulations between languages that share the Latin alphabet, however.
It remains to be investigated whether simulations for other scripts, or indeed between scripts,
are feasible with Multilink \cite[pp. 706--707]{dijkstra2019modelling}.

Unlike the IA and BIA+ models, Multilink does not implement a layer for graphemes or sublexical orthography.
Instead, orthographic representations are activated based on their similarity to the input stimulus.
This abstraction has the advantage of considering characters as singular units,
and therefore the Levenshtein Distance measure may be used to activate orthographic nodes.
We argue that this is a reasonable abstraction for trained readers of the Latin alphabet.
By extension, it seems safe to assume the abstraction works similarly for other alphabets,
like the Greek and Cyrillic alphabet.

For other languages, the implications of such an abstraction of script is less clear.
Consider the Japanese language, which uses not one, but four scripts: the \emph{hiragana} and katakana syllabaries,
\emph{kanji} (Chinese characters), and to a lesser extent \emph{r\=omaji} (Latin characters).
While there is almost certainly confusion between the more complicated \emph{kanji},
we argue we can make an abstraction for the \emph{hiragana} and \emph{katakana}
similar to the one made for alphabets.
Such an abstraction would have positive implications for the generalisability of the model.
Indeed, we hypothesise that, using such an abstraction,
the findings by \shortciteA{miwa2014} could be replicated with Multilink.

\section{Conclusions}

In this thesis, we have discussed the implications of adding lateral inhibition to Multilink.
The parameters regulating lateral inhibition were fit on the basis of reaction times from the
English, British, and Dutch Lexicon Projects \shortcite{balota2007english, keuleers2012british, keuleers2010dutch}.
We find an optimum for these parameters at $OO_\gamma = PP_\gamma = -0.0001$,
giving a maximum correlation of $r = 0.643$ (N=1,205) on these data sets.

Moreover, we have applied Multilink to several smaller experimental studies.
We discussed a neighbourhood study by \shortciteA{mulder2018},
from which stimuli were used as input to the model.
Lateral inhibition was found to improve Multilink's correlations for this study,
yielding an overall correlation $r = 0.67$.
Simulations were also done for the a similar study by \shortciteA{vanlangendonckIP}.
For this study's first experiment, we find an overall correlation $r = 0.69$.
Thus, Multilink provides accurate predictions for this setting.
However, turning to the study's second experiment, with mixed conditions, was less fruitful.
We find Multilink is only able to partially account for the data,
with the resulting correlation $r = 0.44$.

An important part of this thesis dealt with the implications of the study by \citeA{goertz2018},
which involved a demanding combination of cognates and interlingual homographs.
Using the new Word Translation task introduced in this thesis,
We find an overall correlation of 0.538 (N=254) between reaction times and simulated decision times,
as well as a condition pattern correlation of 0.853 (N=8).

In sum, Multilink can now account for not only the \emph{cognate facilitation effect},
but the \emph{interlingual homograph interference effect} as well.
Therefore, Multilink is able to provide an excellent fit to experiment data.

    \cleardoublepage
    \bibliographystyle{apacite}
    \bibliography{bibliography}

    \appendix
    \singlespacing
    \chapter{Multilink Parameters}
\label{app:parameters}

\begin{table}[!ht]
    \centering
    \begin{subtable}{\textwidth}
        \centering
        \begin{tabular}{lp{6cm}}
            \toprule
            Parameter                   & Value \\
            \midrule
            MIN\_ACT                     &  -0.2 \\
            MAX\_ACT                     &  1.0 \\
            DECAY\_RATE                  &  0.07 \\
            \midrule
            MIN\_REST                    &  -0.2 \\
            MAX\_REST                    &  0.0 \\
            MAX\_OPB                     &  0.6402259325203161 \\
            I\_rest                      &  1.0 \\
            O\_rest                      &  -0.2 + OPB $\times$ (0.2 / MAX\_OPB) \\
            L\_rest                      &  -0.2 \\
            S\_rest                      &  -0.2 \\
            P\_rest                      &  -0.2 + OPB $\times$ (0.2 / MAX\_OPB) \\
            \midrule
            IO\_alpha                    &  let score = (MAX\_L - DIST) / MAX\_L \newline
                                            if score $>$ 0.0 \newline
                                            then IO\_multiplier $\times$ score\textsuperscript{3} \newline
                                            else 0.0 \\
            IO\_multiplier               &  0.2 \\
            \midrule
            SS\_multiplier               &  0.0 \\
            \midrule
            criterion\_value             &  0.72 \\
            shortlist\_input\_threshold  &  0.7 \\
            shortlist\_output\_threshold &  0.5 \\
            \midrule
            timestep\_multiplier         &  1.0 \\
            timestep\_adder              &  0.0 \\
            \bottomrule
        \end{tabular}
        \caption{General parameters}
    \end{subtable}
    \par\bigskip
    \begin{subtable}[t]{0.45\textwidth}
        \centering
        \begin{tabular}[t]{ll}
            \toprule
            Parameter  & Value \\
            \midrule
            OP\_alpha   &  0.03 \\
            OS\_alpha   &  0.03 \\
            PO\_alpha   &  0.03 \\
            PS\_alpha   &  0.3 \\
            SO\_alpha   &  0.03 \\
            SP\_alpha   &  0.3 \\
            \midrule
            LO\_alpha   &  0.0 \\
            LP\_alpha   &  0.0 \\
            OL\_alpha   &  0.0 \\
            PL\_alpha   &  0.0 \\
            \bottomrule
        \end{tabular}
        \caption{Alpha connection parameters}
    \end{subtable}
    \begin{subtable}[t]{0.45\textwidth}
        \centering
        \begin{tabular}[t]{ll}
            \toprule
            Parameter  & Value \\
            \midrule
            OO\_gamma   &  -0.001 \\
            PP\_gamma   &  -0.001 \\
            SS\_gamma   &  -0.5 \\
            \midrule
            LL\_gamma   &  0.0 \\
            LO\_gamma   &  0.0 \\
            LP\_gamma   &  0.0 \\
            OL\_gamma   &  0.0 \\
            PL\_gamma   &  0.0 \\
            \bottomrule
        \end{tabular}
        \caption{Gamma connection parameters}
    \end{subtable}
\end{table}

    \chapter{Grid Search Algorithm}
\label{app:gridsearch}

\vspace{-1em}
For compactness and clarity, all debug statements have been omitted from the listing below.

\begin{longlisting}
    \centering
    \singlespacing
\begin{minted}{python}
INIT_WINDOW_MIN = -1.0
INIT_WINDOW_MAX = 0.0
SAMPLE_SIZE = 20
EPSILON = 0.001

def explore(window_min, window_max):
    window_size = INIT_WINDOW_MAX - INIT_WINDOW_MIN
    last_optimum = 0.0
    results_overall = {}

    while True:
        # Create linear space for window.
        window = np.linspace(window_min, window_max, SAMPLE_SIZE)

        # Explore values within this window.
        results = {}
        for x in window:
            (simFile, resFile) = prepareSimulation(x)

            # Only run these settings if we have not done so yet.
            if not os.path.exists(resFile):
                sim = MultilinkSimulation(simFile)
                while True:
                    status = sim.readStatus()
                    if not status:
                        break

            correl = getCorrelation(resFile)
            results[x] = correl
            results_overall[x] = correl

        # Find out which simulation performed best.
        max_x, max_y = None, None
        for (x, y) in results.items():
            if max_y == None or y > max_y:
                max_x, max_y = x, y

        # Stop if we have not improved on the previous optimum.
        if last_optimum > max_y or max_y - last_optimum < EPSILON:
            break

        last_optimum = max_y

        # Narrow window for next iteration
        window_size /= 2
        window_min = max(max_x - (window_size / 2), INIT_WINDOW_MIN)
        window_max = min(max_x + (window_size / 2), INIT_WINDOW_MAX)

    return results_overall
\end{minted}
    \onehalfspacing
\end{longlisting}

    \chapter{Translation Shortlist Implementation}
\label{app:trans_shortlist}

\begin{longlisting}
    \centering
    \singlespacing
\begin{minted}{java}
package net.rekke.bamodel.tasks.shortlist;

// Imports omitted for brevity.

public class TranslationShortlist
{
    private List<Pool> pools;
    private double threshold_value;
    private int last_cycle = -1;
    private List<NodeTuple> shortlist;

    public TranslationShortlist(List<Pool> pools, double threshold_value)
    {
        this.pools = pools;
        this.threshold_value = threshold_value;
        if (threshold_value == 0.0)
            System.out.println("Warning: initialising TranslationShortlist with"
                "threshold_value = 0.0");
    }

    // Boilerplate implementations omitted for brevity.
    public ShortlistIterator getIterator(int cycle);
    private static String getLanguageForNode(Node node);
    private static Node getSemanticsForNode(Node node);

    private void buildShortlist(List<Pool> pools, int cycle)
    {
        if (cycle == last_cycle)
            return;

        last_cycle = cycle;
        shortlist = new ArrayList<NodeTuple>();

        // Build the shortlist using each of the pools.
        System.out.println("Examining shortlist at cycle: " + cycle);
        for (Pool pool : pools)
        {
            for (Node node : pool.getNodes())
            {
                if (node.getActivationHistory()[cycle] < threshold_value)
                    continue;

                Node semantics = getSemanticsForNode(node);
                if (semantics == null)
                    continue;

                String language = getLanguageForNode(node);
                if (language == null)
                    continue;

                // Add an entry for the <node, semantics> pair to the shortlist.
                NodeTuple np = new NodeTuple(node, semantics, language);
                shortlist.add(np);
            }
        }

        if (shortlist.isEmpty())
            return;

        // Order the shortlist by activation, descending.
        shortlist.sort(new Comparator<NodeTuple>() {
            @Override
            public int compare(NodeTuple lhs, NodeTuple rhs) {
                double lhsAct = lhs.getActivationHistory()[cycle];
                double rhsAct = rhs.getActivationHistory()[cycle];
                if (lhsAct > rhsAct)
                    return -1;
                else if (lhsAct == rhsAct)
                    return 0;
                else
                    return 1;
            }
        });
    }

    public Node findNodeWithMatchingSemantics(int cycle, Node targetSemantics)
    {
        buildShortlist(pools, cycle);

        if (shortlist.isEmpty())
            return null;

        for (NodeTuple nt : shortlist)
        {
            if (nt.getSemantics() == targetSemantics.getSymbol())
                return nt.getNode();
        }

        return null;
    }
}
\end{minted}
    \onehalfspacing
\end{longlisting}

    \chapter{Word Translation Implementation}
\label{app:word_translation}

\begin{longlisting}
    \centering
    \singlespacing
\begin{minted}{java}
package net.rekke.bamodel.tasks;

// Imports omitted for brevity.

public class WordTranslation implements Task
{
    private Parameters parameters;

    private String[] inPoolNames;
    private String[] outPoolNames;

    private double timestep_multiplier;
    private double timestep_adder;

    public WordTranslation(Parameters parameters, String[] inPoolNames, String[] outPoolNames)
    throws NetworkBuildException {
        this.parameters = parameters;

        this.timestep_adder = parameters.getTimestep_adder();
        this.timestep_multiplier = parameters.getTimestep_multiplier();

        if (inPoolNames.length != 1 || inPoolNames[0].isEmpty())
            throw new NetworkBuildException("WordTranslation task requires explicitly setting the"
                "input pool.");

        this.inPoolNames = inPoolNames;

        if (outPoolNames.length != 1 || outPoolNames[0].isEmpty())
            throw new NetworkBuildException("WordTranslation task requires explicitly setting the"
                "output pool.");

        this.outPoolNames = outPoolNames;
    }

    public Response applyTask(Network network, Criterion criterion)
    {
        // Input pools. These generally contain all of either orthographical or phonological pools.
        List<Pool> inPools = new ArrayList<Pool>();
        String inputLanguage = network.getLanguage(this.inPoolNames[0]);
        Symbols.NodeType inType = network.getNodeType(this.inPoolNames[0]);
        for (Pool pool : network.getPoolsByType(inType))
            inPools.add(pool);

        // First, determine the input word. We can do this directly through the Criterion, without
        // using a shortlist.
        double inThreshold = parameters.getShortlist_input_threshold();
        TranslationShortlist inShortlist = new TranslationShortlist(inPools, inThreshold);

        // Try to find the critical input node.
        Node critInputNode = null;
        String critInputSemantics = null;
        int critInputCycle = -1;
        for (int cycle = 1; cycle < network.getNCycles(); cycle++)
        {
            ShortlistIterator iterator = inShortlist.getIterator(cycle);
            for (ShortlistIterator it = iterator; it.hasNext(); )
            {
                NodeTuple nt = it.next();

                // TODO: implement a heuristic akin to a hazard rate that skips this check.
                if (nt.getLanguage().equals(inputLanguage))
                {
                    critInputNode = nt.getNode();
                    critInputSemantics = nt.getSemantics();
                    critInputCycle = cycle;
                    break;
                }
            }

            if (critInputNode != null)
                break;
        }

        if (critInputNode == null)
        {
            System.out.println("No critical input node could be determined.");
            return new Response(Symbols.noResponse, "", 0.0, 0.0, null);
        }

        System.out.println("Determined input node as " + critInputNode.getSymbol() + " after " +
            critInputCycle + " cycles");
        System.out.println("Input node has semantics: " + critInputSemantics);

        // Output pools. These generally contain all of either orthographical or phonological pools.
        List<Pool> outPools = new ArrayList<Pool>();
        String outputLanguage = network.getLanguage(this.outPoolNames[0]);
        Symbols.NodeType outType = network.getNodeType(this.outPoolNames[0]);
        for (Pool pool : network.getPoolsByType(outType))
            outPools.add(pool);

        // Create a shortlist object to iterate over output candidates.
        double outThreshold = parameters.getShortlist_output_threshold();
        TranslationShortlist outShortlist = new TranslationShortlist(outPools, outThreshold);

        // Check semantics for output candidates for every cycle.
        Node critOutputNode = null;
        String critOutputSemantics = null;
        int critOutputCycle = -1;

        for (int cycle = critInputCycle; cycle < network.getNCycles(); cycle++)
        {
            ShortlistIterator iterator = outShortlist.getIterator(cycle);
            for (ShortlistIterator it = iterator; it.hasNext(); )
            {
                NodeTuple nt = it.next();

                // TODO: implement a heuristic akin to a hazard rate that makes this check pass.
                if (!nt.getLanguage().equals(outputLanguage))
                    continue;

                // TODO: implement a heuristic akin to a hazard rate that makes this check pass.
                if (!nt.getSemantics().equals(critInputSemantics))
                    continue;

                critOutputNode = nt.getNode();
                critOutputSemantics = nt.getSemantics();
                critOutputCycle = cycle;
                break;
            }

            if (critOutputNode != null)
                break;
        }

        if (critOutputNode == null)
        {
            System.out.println("No critical output node could be determined...");
            return new Response(Symbols.noResponse, "", 0.0, 0.0, null);
        }

        // Interpolate output cycle.
        double[] activations = critOutputNode.getActivationHistory();
        double interpolatedActivationCycle = Interpolator.interpolate(critOutputCycle - 1,
            activations[critOutputCycle - 1], critOutputCycle, activations[critOutputCycle],
            outThreshold);

        System.out.println("Determined output node as " + critOutputNode.getSymbol() + " after " +
            interpolatedActivationCycle + " (" + critOutputCycle + ") cycles");

        // Return 'winner' response.
        CriterionResponse critOutputResp = new CriterionResponse(true, critOutputNode,
            critOutputNode.getSymbol(), critOutputNode.getPool().getLabel(),
            interpolatedActivationCycle, outThreshold, "TranslationShortlist");
        double timestep = timestep_adder + critOutputResp.getResponseTime() * timestep_multiplier;
        return new Response(critOutputResp.getNodeLabel(), critOutputResp.getPoolLabel(),
            critOutputResp.getHighestCriterionValue(), timestep, critOutputResp);
    }

    public void setNewParameters(Parameters parameters)
    {
        this.timestep_adder = parameters.getTimestep_adder();
        this.timestep_multiplier = parameters.getTimestep_multiplier();
    }
}
\end{minted}
    \onehalfspacing
\end{longlisting}

    \cleardoublepage
\end{document}